\documentclass[letterpaper]{article}
\usepackage{aaai19}

\usepackage{times}
\usepackage{helvet}
\usepackage{courier}
\usepackage{url}
\usepackage[shortlabels]{enumitem}
\usepackage{siunitx}
\usepackage{physics}
\usepackage{amssymb}
\usepackage{amsmath}
\usepackage[makeroom]{cancel}
\usepackage{mathrsfs}
\usepackage{amsthm}

\usepackage{bbm}
\usepackage{graphicx}
\usepackage{subfigure}
\usepackage{multirow}
\usepackage[table,xcdraw]{xcolor}

\usepackage[linesnumbered,vlined,ruled]{algorithm2e}
\SetKwProg{Fn}{Function}{}{}
\SetInd{0.1em}{0.5em}
\SetArgSty{textrm}
\SetCommentSty{emph}

\usepackage{tabularx, booktabs}

\theoremstyle{definition}

\newtheorem{thm}{Theorem}

\theoremstyle{definition}

\usepackage{thmtools}

\declaretheoremstyle[%
  spaceabove=0pt,
  spacebelow=0pt,
  headfont=\normalfont\itshape,%
  postheadspace=1em,%
  qed=\qedsymbol%
]{mystyle}

\graphicspath {{figures/}}
\allowdisplaybreaks

\begin{document}

\title{Lifelong Path Planning with Kinematic Constraints\\for Multi-Agent Pickup and Delivery\thanks{Our research was supported by the National Science Foundation (NSF) under grant numbers 1409987, 1724392, 1817189 and 1837779 as well as a gift from Amazon.
}
}

\author{
Hang Ma, Wolfgang H\"onig, T. K. Satish Kumar, Nora Ayanian, Sven Koenig\\
Department of Computer Science\\
University of Southern California\\
\{hangma, whoenig\}@usc.edu, tkskwork@gmail.com, \{ayanian, skoenig\}@usc.edu
}

\maketitle

\begin{abstract}
The Multi-Agent Pickup and Delivery (MAPD) problem models applications where a large number of agents attend to a stream of incoming pickup-and-delivery tasks. Token Passing (TP) is a recent MAPD algorithm that is efficient and effective. We make TP even more efficient and effective by using a novel combinatorial search algorithm, called Safe Interval Path Planning with Reservation Table (SIPPwRT), for single-agent path planning. SIPPwRT uses an advanced data structure that allows for fast updates and lookups of the current paths of all agents in an online setting. The resulting MAPD algorithm TP-SIPPwRT takes kinematic constraints of real robots into account directly during planning, computes continuous agent movements with given velocities that work on non-holonomic robots rather than discrete agent movements with uniform velocity, and is complete for well-formed MAPD instances. We demonstrate its benefits for automated warehouses using both an agent simulator and a standard robot simulator. For example, we demonstrate that it can compute paths for hundreds of agents and thousands of tasks in seconds and is more efficient and effective than existing MAPD algorithms that use a post-processing step to adapt their paths to continuous agent movements with given velocities.
\end{abstract}

\section{Introduction}

In the Multi-Agent Pickup and Delivery (MAPD) problem \cite{MaAAMAS17}, $m$
agents $a_1 \ldots a_m$ attend to a stream of incoming
pickup-and-delivery tasks in a given 2-dimensional 4-neighbor grid with
blocked and unblocked cells of size $L \times L$ each. Agents have to avoid
collisions with each other. A task $\tau_j$ is characterized by a pickup
cell $s_j$ and a delivery cell $g_j$. The task is inserted into the system at
an unknown time. The task set $\mathcal{T}$ contains the unassigned tasks in the system. An agent not executing a task is
called a free agent. It can be assigned any one task $\tau_j \in \mathcal{T}$ at
a time and then has to move from its current cell via cell $s_j$ to cell
$g_j$, implying that it has to move an object from cell $s_j$ to cell $g_j$
and can carry at most one object at a time. Once it arrives at cell $s_j$, it
starts to execute task $\tau_j$ and is called a task agent. Later, once it arrives at
cell $g_j$, it has executed task $\tau_j$ and becomes a free agent
again. A MAPD instance is solved iff all tasks are executed in a bounded
amount of time after they have been inserted into the system. The MAPD problem
models applications such as warehouse robots that
move shelves \cite{kiva}, aircraft towing robots that move planes
\cite{airporttug16}, and office delivery robots that move packages
\cite{DBLP:conf/ijcai/VelosoBCR15}.

Most MAPD algorithms solve the multi-agent pathfinding problem \cite{MaAIMATTERS17} in an inner loop. The multi-agent pathfinding problem is to compute collision-free paths for multiple agents and is NP-hard to solve optimally \cite{YuLav13AAAI,MaAAAI16}. Ways of solving it (and its variants) include reductions to other well-studied combinatorial problems \cite{YuLav13ICRA,erdem2013general,Surynek15} and dedicated algorithms based on search and other techniques \cite{WHCA,ODA,WangB11,PushAndSwap,DBLP:journals/ai/SharonSGF13,EPEJAIR,MStar,DBLP:journals/ai/SharonSFS15,CohenUK16,MaAAMAS16,MaAAAI17,GTAPF}. See \cite{MaWOMPF16,SoCS2017Surv} for complete surveys.

Token Passing (TP) \cite{MaAAMAS17} is a recent MAPD algorithm that is efficient and
effective. It assumes, like many multi-agent pathfinding algorithms, discrete agent movements in the main compass
directions with uniform velocity but can use MAPF-POST \cite{HoenigICAPS16,HoenigIROS16} in
a post-processing step to adapt its paths to continuous forward movements
with given translational velocities and point turns with given
rotational velocities. However, the resulting paths might then not be
effective since planning is oblivious to this transformation. TP needs to
repeatedly plan time-minimal paths for agents that avoid collisions with the
paths of the other agents. We show how TP can be made even more efficient by
using Safe Interval Path Planning with Reservation Table (SIPPwRT), our contribution to improve SIPP \cite{SIPP} for this and many other applications. We also show how TP can be made more general by
letting SIPPwRT directly compute continuous forward movements and point turns
with given velocities. The resulting MAPD algorithm TP-SIPPwRT
guarantees a safety distance between agents and solves all well-formed MAPD
instances.

\section{TP-SIPPwRT}

TP \cite{MaAAMAS17} is a recent MAPD algorithm that assumes discrete agent
movements in the main compass directions with a uniform velocity of typically
one cell per time unit on a grid.  It is similar to Cooperative A* \cite{WHCA}
and can be generalized to a fully distributed MAPD algorithm. We describe TP
very briefly but its important implication for this paper is that agents repeatedly
plan paths for themselves (in Steps TP1 and TP3 below), considering the other
agents as dynamic obstacles that follow their paths and with which collisions
need to be avoided. The agents use space-time A* for this single-agent path
planning.

A set of endpoints is any subset of cells that contains at least all start
cells of agents and all pickup and delivery cells of tasks.  The pickup and
delivery cells are called task endpoints. The other endpoints are called
non-task endpoints. A MAPD instance is well-formed iff the number of tasks is
finite, there are no fewer non-task endpoints than agents, and there exists a
path between any two endpoints that does not pass through other endpoints
\cite{CapVK15,MaAAMAS17}. TP solves all well-formed MAPD instances
\cite{MaAAMAS17}.

TP operates as follows for a given set of endpoints: It uses a token (a
synchronized block of shared memory) that stores the task set and the current
paths, one for each agent. The system repeatedly updates the task set in the
token to contain all unassigned tasks in the system and then
sends the token to some agent that is currently not following a path. The
agent with the token considers all tasks in the task set whose pickup and
delivery cells are different from the end cells of all paths in the token.
\textbf{TP1:} If such tasks exist, then the agent assigns itself that task among these tasks whose pickup cell it can arrive at the earliest, removes the task from the task set, computes two time-minimal paths in the token, one that moves the agent from its current cell to the pickup cell of the task and then one that moves the agent from the pickup cell to the delivery cell of the task, concatenates the two paths into one path, and stores the resulting path.
\textbf{TP2:} If no such tasks exist and the agent is not in the delivery cell of any task in the task set, then it stores the empty path in the token (to wait at its current cell).
\textbf{TP3:} Otherwise, the agent computes and stores a time-minimal path in the token that moves the agent from its current cell to some endpoint that is different from both the delivery cells of all tasks in the task set and from the end cells of all paths in the token. (This rule is necessary to avoid deadlocks.) Each path the agent computes has two properties: (1) It avoids collisions with all other paths in the token; (2) No other paths in the token use its end cell after its end time.
Finally, the agent releases the token, follows its path, and waits at the end cell of the path.

We now show how TP can be made more general by replacing space-time A* with
SIPPwRT, a version of SIPP that computes continuous forward movements and
point turns with given velocities rather than discrete agent movements
in the main compass directions with uniform velocity. We make some simplifying assumptions throughout
this paper even though TP-SIPPwRT and SIPPwRT could easily be generalized
beyond them, mostly because these assumptions are necessary to compare
TP-SIPPwRT against state-of-the-art MAPD algorithms and, as a bonus, make it
easier to explain TP-SIPPwRT: We assume that each agent $a_i$ is a disk with
radius $R_i \leq L/2$ and use its center as its reference point. The \emph{configuration}
of an agent is a pair of its location (cell) and orientation (main compass
direction). Agents always move from the center of their current unblocked cell
to the center of an adjacent unblocked cell via the following available
actions, besides waiting: a point turn $\pi/2$ rads (ninety
degrees) in either clockwise or counterclockwise direction with a given
rotational velocity and a forward movement to the center of the adjacent
cell with a given translational velocity. The agents can accelerate and decelerate infinitely
fast. The paths of two agents are free of collisions iff the interiors of the
agent disks never intersect when they follow their paths.

\section{SIPPwRT}

Space-time A* and SIPP are two versions of A* that both plan time-minimal
paths for agents from their current cells to given goal cells, considering the
other agents as dynamic obstacles that follow their paths and with which
collisions need to be avoided. They both assume discrete agent movements in
the main compass directions with a uniform velocity of typically one cell per
time unit on a grid.  Space-time A* operates on pairs of cells and time steps,
while SIPP groups contiguous time steps during which a cell is not occupied
into safe (time) intervals for that cell and thus operates on pairs of cells
and safe intervals. This affords the A* search of SIPP pruning opportunities
because it is always preferable for an agent to arrive at a cell earlier
during the same safe interval since it can then simply wait at the cell. Thus,
if the A* search of SIPP has already found a path that arrives at some cell at
some time during some safe interval and then discovers a path that
arrives at the same cell at a later time in the same safe
interval, then it can prune the latter path without losing optimality. SIPP
has already been used for robotics applications
\cite{anytimeSIPP,YakovlevA17}. We generalize it to continuous forward
movements and point turns with given velocities in the following,
where a safe interval for a cell is now a maximal contiguous interval during
which the cell is not occupied by dynamic obstacles. Since SIPPwRT, the
resulting version of SIPP, is guaranteed to discover
collision-free paths (like space-time A*) when used as part of TP, TP-SIPPwRT, the resulting
version of TP, continues to solve all well-formed MAPD instances.

\subsection{Reservation Table and Safe Intervals}

SIPP represents the path of each dynamic obstacle as a chronologically ordered
list of cells occupied by the dynamic obstacle, which is not efficient since
SIPP has to iterate through all these lists to calculate all safe intervals of a
given cell. On the other hand, space-time A* maintains a reservation table
that is indexed by a cell and a time step,
which allows for the efficient calculation of all safe intervals of a given
cell.

SIPPwRT improves upon SIPP using a version of a reservation table that handles
continuous agent movements with given velocities and is indexed by a cell. A
reservation table entry of a given cell is a priority queue that contains all reserved
intervals for that cell in increasing order of their lower bounds. A reserved
interval for a cell is a maximal contiguous interval during which the cell is
occupied by some dynamic obstacle. The reservation table allows SIPPwRT to
implement all operations efficiently that are needed by TP-SIPPwRT, namely to
(1) calculate all safe intervals of a given cell; (2) add reservation table
entries after a new path has been calculated; and (3) delete reservation table
entries that refer to irrelevant times in the past in order to keep the
reservation table small.

\noindent\textbf{Function GetSafeIntervals.}
$\emph{GetSafeIntervals}(\emph{cell})$ returns all safe intervals for
cell $\emph{cell}$ in increasing order of their lower bounds. The safe
intervals for the cell are obtained as the complements of the reserved
intervals for the cell with respect to interval $[0, \infty]$. For safe
interval $i = [i.\emph{lb}, i.\emph{ub}]$ and a dynamic obstacle departing from cell $\emph{cell}$ at time $i.\emph{lb}$, $\emph{dep\_cfg}[\emph{cell},i]$ is the configuration of
the dynamic obstacle at time $i.\emph{lb}$. It is $\emph{NULL}$ iff $i.\emph{lb} \leq \emph{current\_t}$. Similarly, for safe interval $i = [i.\emph{lb}, i.\emph{ub}]$ and the dynamic obstacle arriving at cell $\emph{cell}$ at time $i.\emph{ub}$, $\emph{arr\_cfg}[\emph{cell},i]$ is the configuration of the dynamic obstacle at time $i.\emph{ub}$. It is $\emph{NULL}$ iff $i.\emph{ub} = \infty$.

\subsection{Time Offsets}

The safe intervals of a cell represent the times during which the cell is not
occupied. However, this does not mean that an agent can arrive at any of
those times at the cell since the agent might still collide with a dynamic
obstacle that has just departed from the cell or is about to arrive at the
cell. Thus, the lower and upper bounds of a safe interval have to be tightened using the following time offsets.

Function \emph{Offset}(\emph{cfg1}, \emph{cfg2}) returns the time offset
$\Delta T$ that expresses the minimum amount of time the center of some unknown
agent $a_1$ with safety radius $R_1$ and translational velocity
$v_{\emph{trans,1}}$ needs to depart from the center of a cell $\emph{cfg1}.\emph{cell} =
l$ with configuration $\emph{cfg1}$ before the center of some unknown
agent $a_2$ with safety radius $R_2$ and translational velocity
$v_{\emph{trans,2}}$, coming from some cell $l'$, arrives at the center of
the same cell $\emph{cfg2}.\emph{cell} = l$ with configuration $\emph{cfg2}$ to avoid a collision. The time offset is zero iff either $\emph{cfg1} = \emph{NULL}$ or $\emph{cfg2} =
\emph{NULL}$, meaning that either agent $a_1$ or agent $a_2$ does not exist. The
calculation of the time offset requires only knowledge of the
configurations, safety radii, and velocities of both agents.\footnote{In the pseudocode of SIPPwRT, we show how to keep track of the configurations
  but do not include the safety radii and velocities in the configurations
  for ease of readability (although this needs to be done in case they are not the same for all agents and times).}

Assume that agent $a_2$ departs from cell $l'$ at time 0 (and thus arrives at
cell $l$ at time $t' = \frac{L}{v_{\emph{trans,2}}}$) and agent $a_1$ departs
from cell $l$ at time $t_d \leq \frac{L}{v_{\emph{trans,2}}}$. $D(t)$ is the distance between
the agents, where $t$ is the amount of time elapsed after agent $a_2$ departs
from cell $l'$. It must hold that $D(t) \ge R_1 + R_2$ to avoid that the two agents collide.  We distinguish three cases to calculate the time offset $\Delta T$:

\subsubsection{(a) Same Direction.}

\begin{figure}[t]
  \centering
  \includegraphics[width=0.324\columnwidth]{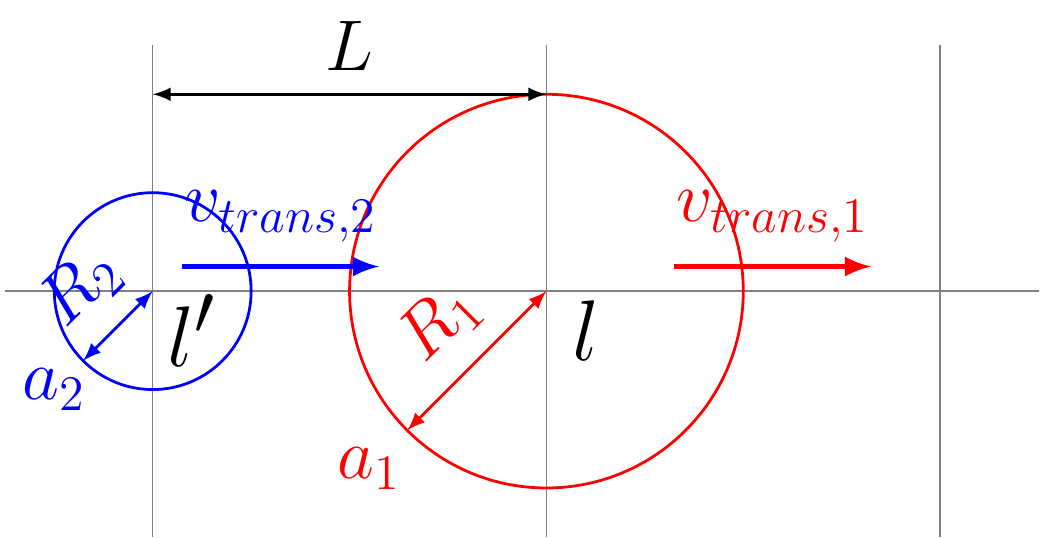}\hspace{.5ex}\includegraphics[width=0.324\columnwidth]{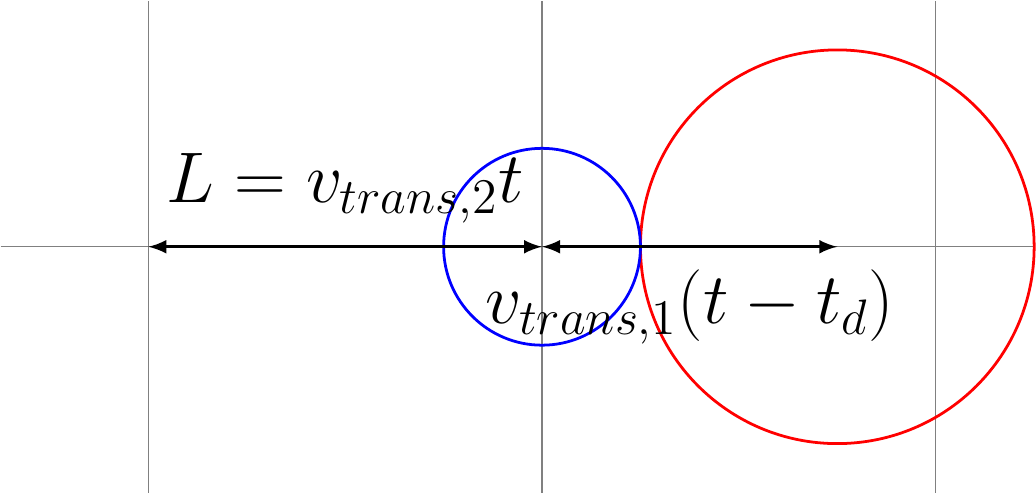}\hspace{.5ex}\includegraphics[width=0.324\columnwidth]{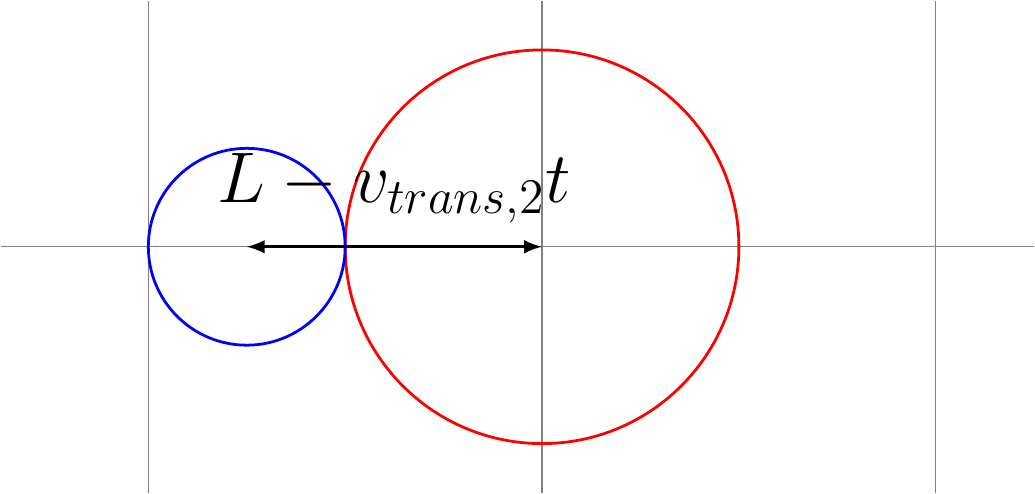}
  \caption{Left: Two agents move in the same direction. Middle: $D$ is at its
    minimum for the $v_{\emph{trans,1}}<v_{\emph{trans,2}}$ case. Right: $D$ is at its
    minimum for the $v_{\emph{trans,1}} \geq v_{\emph{trans,2}}$ case.}\label{fig:horizontal}
\end{figure}

Both agents move in the same direction (meaning that the orientations of
configurations \emph{cfg1} and \emph{cfg2} are equal), see Figure \ref{fig:horizontal}
(left), where gray lines connect the centers of cells. In this case, $D(t) = L- v_{\emph{trans,2}} t + v_{\emph{trans,1}} (t
- t_d)$. We now distinguish two sub-cases to show that the time offset is
$\Delta T = \frac{R_1 + R_2}{\min(v_{\emph{trans,1}}, v_{\emph{trans,2}})}$.

\textbf{(a1) Case $v_{\emph{trans,1}} < v_{\emph{trans,2}}$.} This case is
shown in Figure \ref{fig:horizontal} (middle). $D(t)$ decreases as the time
$t$ increases. $D(t)$ thus reaches its minimum at the time $t = t'' =
\frac{L}{v_{\emph{trans,2}}}$ when agent $a_2$ arrives at cell
$l$. Substituting $t = \frac{L}{v_{\emph{trans,2}}}$ back into $D(t) \ge R_1 +
R_2$, we have {\scriptsize
\begin{align*}
  D(t) &= L- v_{\emph{trans,2}}t + v_{\emph{trans,1}}(t - t_d) = v_{\emph{trans,1}}(\frac{L}{v_{\emph{trans,2}}} - t_d) &\ge R_1 + R_2.
\end{align*}
}
Therefore, $t_d \le \frac{L}{v_{\emph{trans,2}}} - \frac{R_1 + R_2}{v_{\emph{trans,1}}}$.  The time offset
$\Delta T$ is thus $\Delta T = t' - \max t_d = \frac{L}{v_{\emph{trans,2}}} - (\frac{L}{v_{\emph{trans,2}}}
- \frac{R_1 + R_2}{v_{\emph{trans,1}}}) = \frac{R_1 + R_2}{v_{\emph{trans,1}}}$.

\textbf{(a2) Case $v_{\emph{trans,1}} \geq v_{\emph{trans,2}}$.} This case is shown in Figure
\ref{fig:horizontal} (right). $D(t)$ decreases before agent $a_1$ starts to
move and then increases as the time $t$ increases. $D(t)$ thus reaches its
minimum at the time $t = t_d$ when agent $a_1$ starts to move. Substituting $t
= t_d$ back into $D(t) \ge R_1 + R_2$, we have
{\scriptsize
\begin{align*}
  D(t) &= L- v_{\emph{trans,2}} t + v_{\emph{trans,1}}(t - t_d) = L-
  v_{\emph{trans,2}} t_d &\ge R_1 + R_2.
\end{align*}
}
Therefore, $t_d \leq \frac{L- (R_1 + R_2)}{v_{\emph{trans,2}}}$.  The time offset is thus
$\Delta T = t' - \max t_d = \frac{L}{v_{\emph{trans,2}}} - \frac{L- (R_1 + R_2)}{v_{\emph{trans,2}}} =
\frac{R_1 + R_2}{v_{\emph{trans,2}}}$.

\subsubsection{(b) Orthogonal Directions.}

\begin{figure}[t]
  \centering
  \includegraphics[width=0.216\columnwidth]{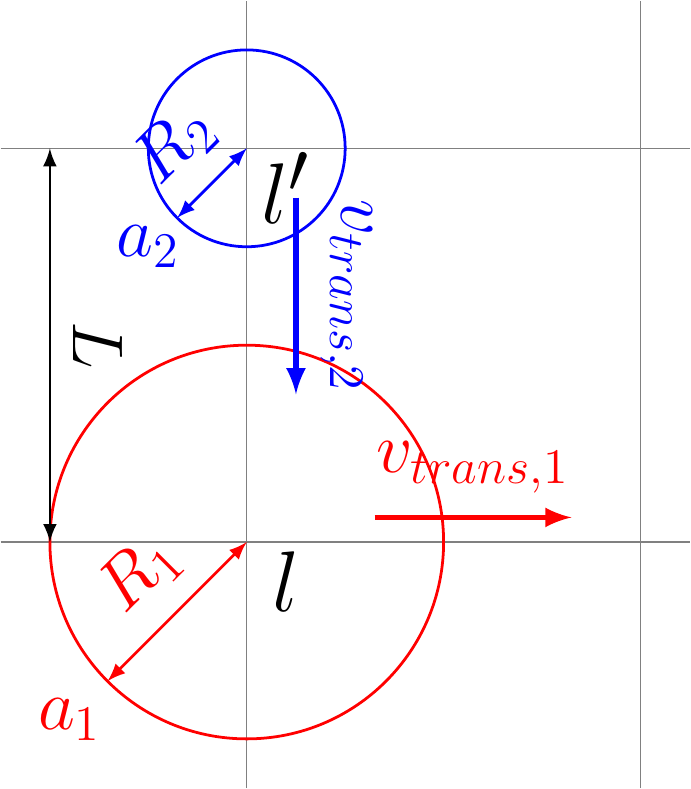}\hspace{1ex}
  \includegraphics[width=0.216\columnwidth]{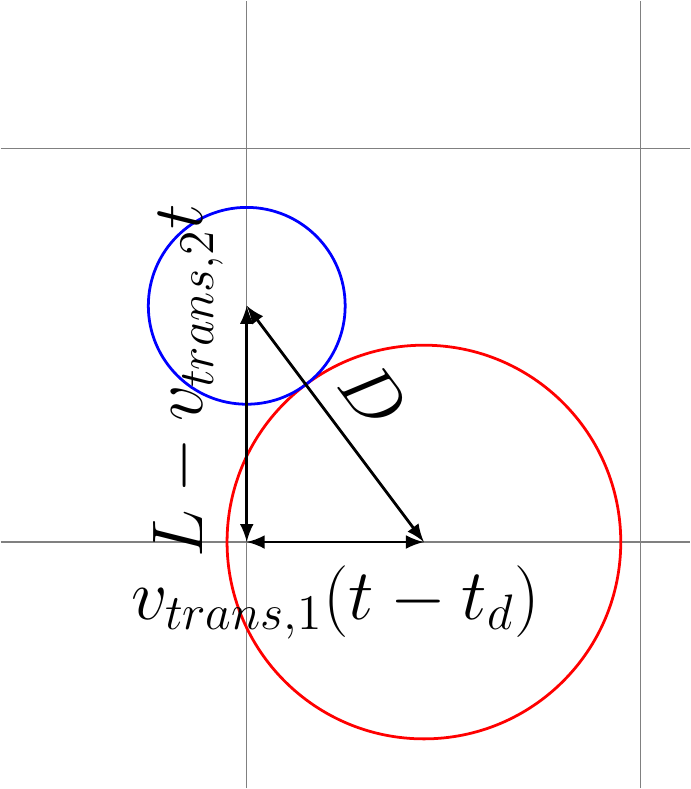}
  \caption{Left: Two agents move in orthogonal directions. Right: $D$ is at its minimum.}\label{fig:orthogonal}
\end{figure}

Both agents move in orthogonal directions, see Figure \ref{fig:orthogonal}. In
this case, $D(t)$ $=$ $\sqrt{(v_{\emph{trans,1}}(t - t_d))^2 + (L-
  v_{\emph{trans,2}} t)^2}$. We determine the
time $t$ at which $D(t) \geq 0$ reaches its minimum by solving $\pdv{D^2}{t} =
0$. Substituting the result $t = \frac{v_{\emph{trans,1}}^2t_d +
  v_{\emph{trans,2}} L}{v_{\emph{trans,1}}^2 + v_{\emph{trans,2}}^2}$ into
$D^2 \ge (R_1 + R_2)^2$, we have
{\scriptsize
\begin{align*}
 D^2(t) &= (L- v_{\emph{trans,2}}t)^2 + (v_{\emph{trans,1}}(t - t_d))^2 \\ &= \frac{(v_{\emph{trans,1}}(v_{\emph{trans,2}}t_d - L))^2}{(v_{\emph{trans,1}}^2 + v_{\emph{trans,2}}^2)} \ge (R_1 + R_2)^2.
\end{align*}
}
Since $L\ge v_{\emph{trans,2}}t_d$, we have $v_{\emph{trans,1}}(L- v_{\emph{trans,2}}t_d)$ $\ge$ $\sqrt{v_{\emph{trans,1}}^2 + v_{\emph{trans,2}}^2}(R_1 +
R_2)$.  Therefore, $t_d$ $\le$ $\frac{v_{\emph{trans,1}}L- \sqrt{v_{\emph{trans,1}}^2 + v_{\emph{trans,2}}^2}(R_1 +
  R_2)}{v_{\emph{trans,1}}v_{\emph{trans,2}}}$.  The time offset is thus $\Delta T$ $=$ $t' - \max t_d$ $=$
$\frac{L}{v_{\emph{trans,2}}}$ $-$ $\frac{v_{\emph{trans,1}} L- \sqrt{v_{\emph{trans,1}}^2 + v_{\emph{trans,2}}^2}(R_1 + R_2)}{v_{\emph{trans,1}}v_{\emph{trans,2}}} =
\frac{\sqrt{v_{\emph{trans,1}}^2 + v_{\emph{trans,2}}^2}(R_1 + R_2)}{v_{\emph{trans,1}}v_{\emph{trans,2}}}$.

\subsubsection{(c) Opposite Directions.}

Both agents move in opposite directions, that is, agent $a_1$ moves from cell
$l$ to cell $l'$ and agent $a_2$ moves from cell $l'$ to cell $l$. The time
offset is set to allow agent $a_1$ to arrive at cell $l'$ even before agent $a_2$ departs from cell $l'$. In this case, the time offset is $\Delta T =
\frac{L}{v_{\emph{trans,1}}} + \frac{L}{v_{\emph{trans,2}}}$, which is the sum
of the times that agent $a_1$ needs to move from cell $l$ to cell $l'$ and
that agent $a_2$ needs to move from cell $l'$ to cell $l$. We later show that SIPPwRT avoids collisions when it uses all bounds simultaneously.

\subsection{Increased/Decreased Bounds}

The algorithm calls the following functions to tighten the lower and upper bounds of safe interval $i$ during which an agent can stay at cell $l = \emph{cfg}.\emph{cell}$ safely.

The algorithm calls Function $\emph{GetLB1}(\emph{cfg}, i)$ for an agent $a_2$ to return $\max_{j} (j.\emph{lb} + \emph{Offset}(\emph{dep\_cfg}[\emph{cfg}.\emph{cell},j],\emph{cfg}))$.
Here, $j.\emph{lb} + \emph{Offset}(\emph{dep\_cfg}[\emph{cfg}.\emph{cell},j],\emph{cfg})$ is the increased lower bound for each safe interval $j$ in $\emph{GetSafeIntervals}(\emph{cfg}.\emph{cell})$ with $j.\emph{lb} \leq i.\emph{lb}$. For agent $a_2$ that arrives at cell $l = \emph{cfg}.\emph{cell}$ from another cell $l'$ with configuration $\emph{cfg}$, the idea is to prevent it from colliding with any dynamic obstacle $a_1$ that departs from cell $l$ with configuration $\emph{dep\_cfg}[\emph{cfg}.\emph{cell},j]$ before $a_2$ arrives at cell $l$.

The algorithm calls Function $\emph{GetUB1}(\emph{cfg}, i)$ for an agent $a_1$ to return $\min_{j} (j.\emph{ub} - \emph{Offset}(\emph{cfg}, \emph{arr\_cfg}[\emph{cfg}.\emph{cell},j]))$. Here, $j.\emph{ub} - \emph{Offset}(\emph{cfg},\emph{arr\_cfg}[\emph{cfg}.\emph{cell},j])$ is the decreased upper bound for each safe interval $j$ in $\emph{GetSafeIntervals}(\emph{cfg}.\emph{cell})$ with $j.\emph{ub} \geq i.\emph{ub}$. For agent $a_1$ that departs from cell $l=\emph{cfg}.\emph{cell}$ with configuration $\emph{cfg}$, the idea is to prevent it from colliding with any dynamic obstacle $a_2$ that arrives at cell $l$ from another cell $l'$ with configuration $\emph{arr\_cfg}[\emph{cfg}.\emph{cell},j]$ after $a_1$ departs from cell $l$.

The algorithm calls Function $\emph{GetLB2}(\emph{cfg}, i)$ for an agent $a_1$ to return $\max_{j}(j.\emph{lb} + \frac{L}{v_\emph{trans}'} - \frac{L}{v_\emph{trans}})$. Here, $j.\emph{lb} + \frac{L}{v_\emph{trans}'} - \frac{L}{v_\emph{trans}}$ is the increased lower bound for each safe interval $j$ in $\emph{GetSafeIntervals}(\emph{cfg}.\emph{cell})$ where the orientation of $\emph{dep\_cfg}[\emph{cfg}.\emph{cell},j]$ is the same as that of $\emph{cfg}$ and $j.\emph{lb} \leq i.\emph{lb}$. For agent $a_1$ that departs from cell $l = \emph{cfg}.\emph{cell}$ with configuration $\emph{cfg}$ and translational velocity $v_{\emph{trans}}$ and moves also toward cell $l'$, the idea is to prevent it from arriving at cell $l'$ earlier than (and thus ``passing through'') any dynamic obstacle $a_2$ that departs from cell $l$ before agent $a_1$ with configuration $\emph{dep\_cfg}[\emph{cfg}.\emph{cell},j]$ and translational velocity $v_{\emph{trans}}'$ and moves also toward cell $l'$.

The algorithm calls Function $\emph{GetUB2}(\emph{cfg}, i)$ for an agent $a_1$ to return $\min_{j}(j.\emph{lb} + \frac{L}{v_\emph{trans}'} - \frac{L}{v_\emph{trans}})$. Here, $j.\emph{lb} + \frac{L}{v_\emph{trans}'} - \frac{L}{v_\emph{trans}}$ is the decreased upper bound for each safe interval $j$ in $\emph{GetSafeIntervals}(\emph{cfg}.\emph{cell})$ where the orientation of $\emph{dep\_cfg}[\emph{cfg}.\emph{cell},j]$ is the same as that of $\emph{cfg}$ and $j.\emph{lb} \geq i.\emph{ub}$. For agent $a_1$ that departs from cell $l = \emph{cfg}.\emph{cell}$ with configuration $\emph{cfg}$ and translational velocity $v_{\emph{trans}}$ and moves also toward cell $l'$, the idea is to prevent it from arriving at cell $l'$ later than (and thus ``being passed through'' by) any dynamic obstacle $a_2$ that departs from cell $l$ after agent $a_1$ with configuration $\emph{dep\_cfg}[\emph{cfg}.\emph{cell},j]$ and translational velocity $v_{\emph{trans}}'$ and moves toward cell $l'$.

\subsection{Admissible H-Values}

Step TP1 of TP-SIPPwRT requires an agent to use SIPPwRT twice, namely (1) to plan a time-minimal path from its current configuration to a candidate set of
endpoints (pickup cells) and (2) to plan a time-minimal path from the resulting configuration to a particular endpoint (a delivery cell). Step
TP3 requires the agent to use SIPPwRT once to plan a time-minimal path from
its current configuration to a candidate set of endpoints (to avoid deadlocks). The agent always moves
along each path with given (fixed) translational and rotational
velocities (unless it waits). Thus, SIPPwRT has to plan only paths to a given
set $G$ of one or more endpoints. By ignoring the dynamic obstacles, we determine the admissible h-values needed
for the A* search of SIPPwRT to plan time-minimal paths as follows: We
calculate a time-minimal path that excludes waiting for the agent from each
configuration $\emph{cfg}$ to each configuration $\emph{cfg}'$ whose cell is an endpoint (by
searching backward once from each configuration $\emph{cfg}'$). We then use the
minimum heuristic \cite{SternGF17} $h(\emph{cfg},G) = \min_{\emph{cfg}'.\emph{cell}\in G} h(\emph{cfg},\emph{cfg}')$ as
admissible h-value of configuration $\emph{cfg}$, where $h(\emph{cfg},\emph{cfg}')$ is the
calculated time of the time-minimal path from $\emph{cfg}$ to $\emph{cfg}'$. In practice, if
set $G$ is large and endpoints are densely distributed across the grid, it is
more efficient to use $h(\emph{cfg},G) = 0$ (as we do for Step TP3 of TP-SIPPwRT)
since it can be calculated faster even though SIPPwRT might expand more nodes.

\subsection{Pseudocode}

Algorithm \ref{alg:SIPPwRT} shows the pseudocode of SIPPwRT, which plans a
time-minimal path for an agent with translational velocity $v_{\emph{trans}}$
and rotational velocity $v_{\emph{rot}}$ from its configuration
\emph{start\_cfg} at time $\emph{current\_t}$ to a cell in set $G$. SIPPwRT
performs a regular A* search with nodes that are pairs of configurations of
the agent and safe intervals. The g-value $g[n]$ of a node $n = \langle n.\emph{cfg},
n.\emph{int} \rangle$ with configuration $n.\emph{cfg}$ and safe interval $n.\emph{int} =
[n.\emph{int}.\emph{lb}, n.\emph{int}.\emph{ub}]$ is the earliest discovered time in $n.\emph{int}$
when the agent can be in configuration $n.\emph{cfg}$.
The start node is $n=\langle
\emph{start\_cfg} , [\emph{current\_t},\infty] \rangle$ with $g[n] =
\emph{current\_t}$. The safe interval $n.\emph{int}$ of the start node
expresses that the agent can wait forever in its current configuration. A node $n$ is a goal node iff the cell of its configuration is in set $G$ and the agent can wait forever in its configuration ($n.\emph{int}.\emph{ub} = \infty$).

In our implementation of SIPPwRT, each action is a \emph{turn-and-move} action, i.e., a point turn into one of the four compass directions followed by a wait (when necessary) and then a forward movement to a neighboring unblocked cell. Since only forward movements define the temporal constraints between safe intervals of neighboring cells, the state space of our search remains unaffected by the use of turn-and-move actions instead of separate point turn, move, and wait actions independently.

\begin{algorithm}[t!]
\setlength\hsize{8cm}
\scriptsize
\linespread{0}\selectfont
\caption{\scriptsize SIPPwRT.}
\label{alg:SIPPwRT}
\Fn{SIPPwRT(\text{start\_cfg}, $G$, \text{current\_t}, $v_{\text{trans}}$,
  $v_{\text{rot}}$)}
   {
     $n_{\emph{start}} \gets \emph{NewNode}(\langle\emph{start\_cfg},
     [\emph{current\_t}, \infty]\rangle$)\label{lin:start_node}\;
     $g[n_{\emph{start}}]\ \gets \emph{current\_t}$\label{lin:start_g_value}\;
     $\emph{OPEN} \gets \{n_{\emph{start}}\}$\label{lin:insert_start}\;
     \While{$\emph{OPEN}\neq \emptyset$\label{lin:open_not_empty}}
           {
             $n \gets \arg \min_{n' \in \emph{OPEN}} (g[n'] + h(n'.\emph{cfg},G))$\label{lin:best_node}\;
             $\emph{OPEN} \gets \emph{OPEN} \setminus \{ n \}$\label{lin:remove_best_node}\;
             \If{$n.\emph{cell} \in G$ and $n.\emph{int}.\emph{ub}=\infty$\label{lin:is_goal}}
                {
                  \Return path from \emph{start\_cfg} to $n.\emph{cfg}$\label{lin:return_path}\;
                }
                $\emph{successors} \gets \emph{GetSuccessors}(n)$\label{lin:all_successors}\;
                \ForEach{$n'\in \emph{successors}$\label{lin:each_successor}}
                        {
                          \If{$g[n']$ is undefined\label{lin:is_undefined}}
                             {
                               $g[n'] \gets \infty$\label{lin:init_g_value}\;
                             }
                             \If{$g[n'] > g[n] + \emph{cost}[n, n']$\label{lin:smaller_g_value}}
                                {
                                  $\emph{parent}[n']\gets n$\label{lin:update_parent}\;
                                  $g[n'] \gets g[n] + \emph{cost}[n, n']$\label{lin:update_g_value}\;
                                  \If{$n'\notin \emph{OPEN}$\label{lin:not_in_open}}
                                  {
                                    $\emph{OPEN}\gets\emph{OPEN}\cup\{n'\}$\label{lin:insert_node}\;
                                  }
                                }
                        }
           }
           \Return no path exists (does not happen for well-formed MAPD
           instances)\;
   }
\Fn{GetSuccessors($n$)}
{
   $\emph{successors} \gets \emptyset$\;
   \ForEach{legal turn-and-move $\emph{action}$ in $n$\label{lin:legal_actions}}
   {
     $\emph{cfg\_t} \gets$ configuration resulting from executing the point turn of \emph{action} in $n.\emph{cfg}$ ($\emph{cfg\_t}.\emph{cell} = n.\emph{cfg}.\emph{cell}$)\label{lin:turned_cfg}\;
     $\emph{ub1}\gets \emph{GetUB1}(\emph{cfg\_t}, n.\emph{int})$\label{lin:ub1}\;
     $\emph{lb2} \gets \emph{GetLB2}(\emph{cfg\_t}, n.\emph{int})$\label{lin:lb2}\;
     $\emph{ub2} \gets \emph{GetUB2}(\emph{cfg\_t}, n.\emph{int})$\label{lin:ub2}\;
     $\emph{lb}\gets\max( (g[n] + \Delta t_\emph{turn}(\emph{action}, v_{\emph{rot}})), \emph{lb2})$\label{lin:increased_lb}\;
     $\emph{ub}\gets\min( \emph{ub1}, \emph{ub2})$\label{lin:decreased_ub}\;
     \If{$\emph{lb} \leq \emph{ub}$\label{lin:can_depart}}
     {
       $\emph{cfg}' \gets$ configuration resulting from executing the forward movement in \emph{action} in $n.\emph{cfg\_t}$\label{lin:new_cfg}\;
       $i'.\emph{lb}\gets \emph{lb} + \Delta t_\emph{move}(\emph{action}, v_{\emph{trans}})$\label{lin:arrive_lb}\;
       $i'.\emph{ub}\gets \emph{ub} + \Delta t_\emph{move}(\emph{action}, v_{\emph{trans}})$\label{lin:arrive_ub}\;
       $\emph{safeIntervals} \gets \emph{GetSafeIntervals}(\emph{cfg}'.\emph{cell})$\label{lin:all_safe_int}\;
       \ForEach{$i'' \in \emph{safeIntervals}$\label{lin:each_safe_int}}
       {
         $\emph{lb1} \gets \emph{GetLB1}(\emph{cfg}', i'')$\label{lin:lb1}\;
         \If{$[\emph{lb1}, i''.\emph{ub}] \cap i' \neq\emptyset$\label{lin:interval_intersection}}
         {
           $t' \gets \max(i'.\emph{lb}, \emph{lb1})$\label{lin:earliest_arrival}\;
           $n'\gets \emph{NewNode}(\langle\emph{cfg}', i''\rangle)$\label{lin:new_succ}\;
           $\emph{cost}[n,n']\gets t' - g[n]$\label{lin:cost_with_wait}\;
           $\emph{successors}\gets\emph{successors}\cup\{n'\}$\label{lin:insert_new_succ}\;
         }
       }
     }
  }
  \Return \emph{successors}\;
}
\end{algorithm}

\noindent\textbf{Function GetSuccessors.}
$\emph{GetSuccessors}(n)$ calculates the successors of node $n$ by
considering all legal turn-and-move actions $\emph{action}$ available to the agent in
configuration $n.\emph{cfg}$ [Line \ref{lin:legal_actions}]. Assume that executing the point turn of action $\emph{action}$ takes $\Delta t_\emph{turn}(\emph{action}, v_\emph{rot})$ time units and results in configuration $\emph{cfg\_t}$ with which the agent departs from its current cell [Line \ref{lin:turned_cfg}]. The agent must depart from its current cell no later than $\emph{lb}$ and no earlier than $\emph{ub}$ to avoid colliding with dynamic obstacles that also visit its current cell [Lines \ref{lin:ub1}-\ref{lin:decreased_ub}].
If the agent can depart from its current cell [Line \ref{lin:can_depart}], then assume that executing the forward movement of action $\emph{action}$ in
configuration $\emph{cfg\_t}$ takes $\Delta t_\emph{move}(\emph{action}, v_\emph{trans})$ time units and results in successor configuration $\emph{cfg}'$
[Line \ref{lin:new_cfg}]. The agent waits an appropriate amount of time in configuration $\emph{cfg\_t}$ after the point turn, then executes the forward movement, and arrives in configuration $\emph{cfg}'$ in
interval $i' = [\emph{lb}+\Delta t_\emph{move}(\emph{action}, v_\emph{trans}), \emph{ub}+\Delta t_\emph{move}(\emph{action}, v_\emph{trans})]$ [Lines \ref{lin:arrive_lb}-\ref{lin:arrive_ub}].
The successors of node $n$ are generated by processing all safe intervals $i'' = [i''.\emph{lb},i''.\emph{ub}]$ for the new
cell $\emph{cfg}'.\emph{cell}$ of the agent [Lines \ref{lin:all_safe_int}-\ref{lin:each_safe_int}]. The lower bound of safe interval
$i''$ is increased from $i''.\emph{lb}$ to $\emph{lb1}$ to ensure that the agent
can arrive at its new cell without colliding with dynamic obstacles that
also visit its new cell [Line \ref{lin:lb1}]. The updated safe interval $[\emph{lb1}, i''.\emph{ub}]$ is intersected with interval $i'$ [Line \ref{lin:interval_intersection}]. If their intersection is
non-empty, then the agent can arrive at its successor configuration during
safe interval $i''$. Only the earliest time $t'$ in the intersection needs to
be considered (since the agent can simply wait in its successor configuration
and the later times in the intersection can thus be pruned, as argued earlier)
[Line \ref{lin:earliest_arrival}].  The resulting successor of node $n$ is $n' = \langle \emph{cfg}', i''
\rangle$ [Line \ref{lin:new_succ}], and the cost (here: time) of the transition from node $n$
to node $n'$ is $cost[n,n'] = t'-g[n]$ [Line \ref{lin:cost_with_wait}] (consisting of executing the point turn of action $\emph{action}$ for $\Delta t_\emph{turn}(\emph{action}, v_\emph{rot})$ time units, waiting for
$t'-g[n]-\Delta t_\emph{turn}(\emph{action}, v_\emph{rot})-\Delta t_\emph{move}(\emph{action}, v_\emph{trans})$ time units, and then executing the forward movement of action $\emph{action}$ for $\Delta t_\emph{move}(\emph{action}, v_\emph{trans})$
time units), so that $g[n'] = g[n] + cost[n,n'] = t'$ is the earliest
discovered time in $n'.\emph{int} = i''$ when the agent can be in configuration
$n'.\emph{cfg} = \emph{cfg}'$.

\noindent\textbf{Main Routine.}
The main routine of SIPPwRT performs a regular A* search. It initializes the
g-value of the start node and inserts the node into the OPEN list [Lines
  \ref{lin:start_node}-\ref{lin:insert_start}]. It then repeatedly removes a node $n$ with the smallest sum of
g-value and h-value $g[n] + h(n.\emph{cfg}, G)$ from the OPEN list [Lines \ref{lin:best_node}-\ref{lin:remove_best_node}] and
processes it: If the node is a goal node, then it returns the path found by
following the parent pointers from the node to the start node [Lines
  \ref{lin:is_goal}-\ref{lin:return_path}]. Otherwise, it generates the successors of the node [Line \ref{lin:all_successors}]. For each
successor, it initializes its g-value to infinity if the g-value is still
undefined [Lines \ref{lin:is_undefined}-\ref{lin:init_g_value}]. It then checks whether the g-value of the successor
can be reduced by changing its parent pointer to node $n$ [Line \ref{lin:smaller_g_value}]. If so, it
changes the parent pointer of the successor, reduces its g-value, and inserts
it into the OPEN list (if necessary) [Lines \ref{lin:update_parent}-\ref{lin:insert_node}].


\begin{thm}
The path returned by SIPPwRT from the start configuration to a goal is free of collisions.
\end{thm}
We prove Theorem 1 in the technical report.
Since all heuristics used by SIPPwRT are admissible as argued earlier, using the argument in \cite{SIPP} together with Theorem 1, it is straightforward to show that SIPPwRT returns a time-minimal path to a given set $G$ of one or more endpoints that does not collide with the paths of other agents in the token and is complete for the single-agent path-planning problems for function calls TP1 and TP3. We can thus rely on the proof of Theorem 3 in \cite{MaAAMAS17} to show that TP-SIPPwRT is complete for well-formed MAPD instances.

\begin{thm}\label{thm2}
TP-SIPPwRT solves all well-formed MAPD instances.
\end{thm}

\section{Simulated Automated Warehouses}

\begin{figure}[t]
  \centering
  \includegraphics[height=40pt]{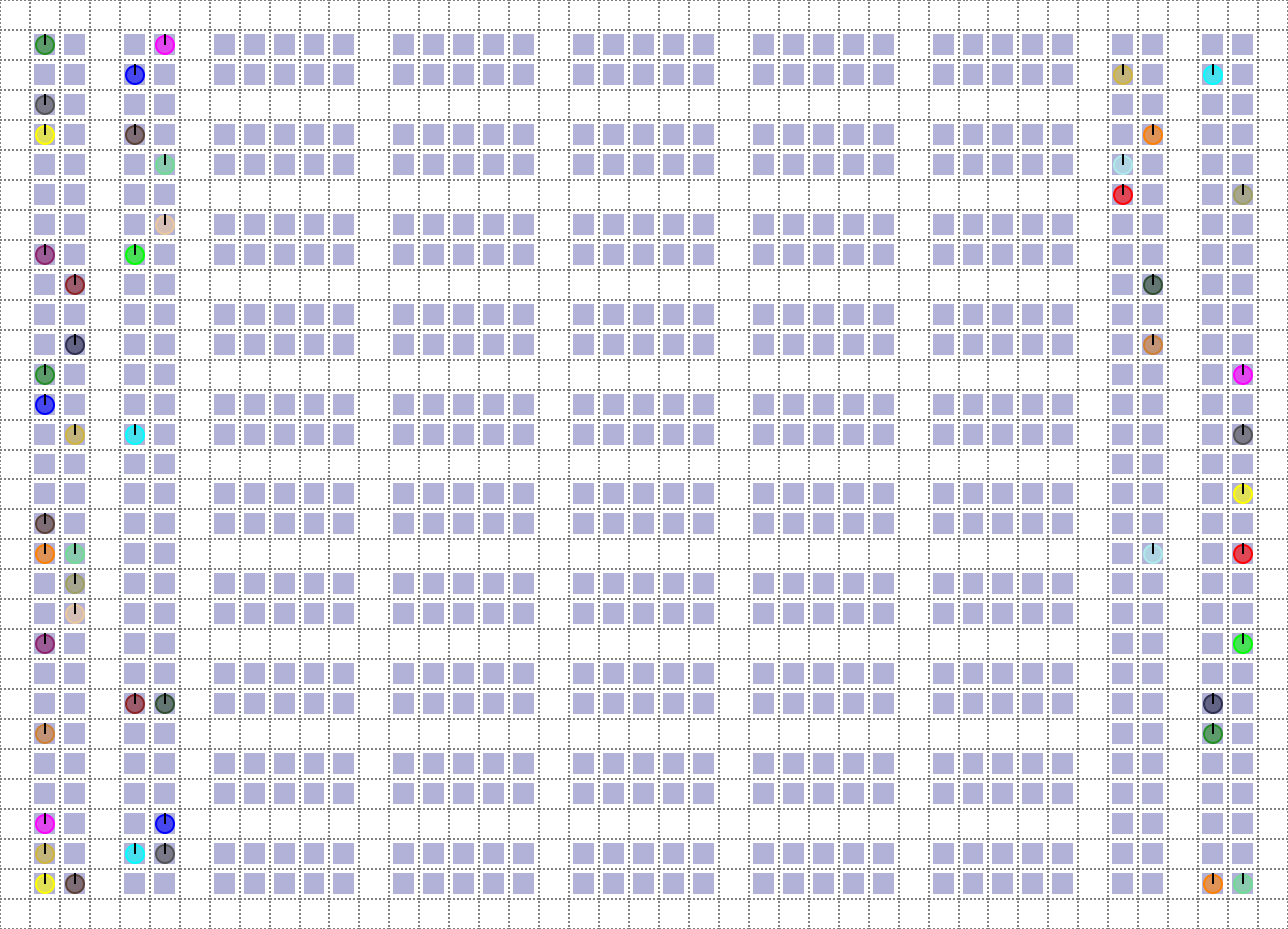}\hspace{.5ex}\includegraphics[height=40pt]{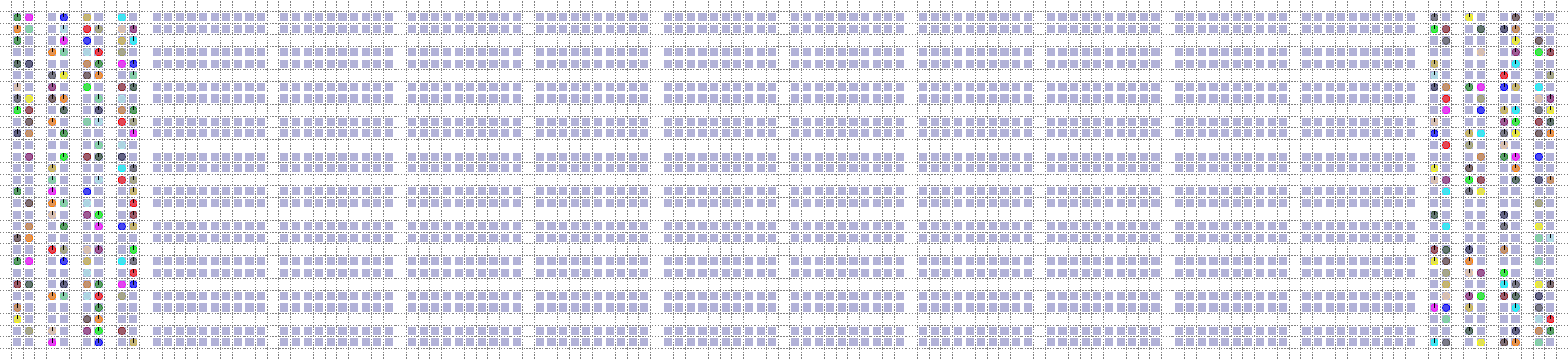}
  \caption{Left: Small simulated warehouse environment. Right: Large simulated warehouse environment.}\label{fig:warehouse}
\end{figure}

We demonstrate the benefits of TP-SIPPwRT for automated warehouses using both
an agent simulator with perfect path execution and a standard robot simulator
with imperfect path execution resulting from unmodeled kinodynamic constraints
and motion noise by the MAPD algorithms. Figure \ref{fig:warehouse} (left)
shows an example on the agent simulator with 50 agents and cells of size \SI{1
  x 1}{m}. Grey cells in columns of grey cells are potential start cells for
the agents. Colored disks are the actual start cells, which are drawn randomly
from all potential start cells and are the non-task endpoints. All agents face
north in their start cells. Grey cells other than the start cells are task
endpoints (that would house shelves in a warehouse even though we do not model
shelves here). The pickup and delivery cells of all tasks are drawn randomly
from all task endpoints. White cells are non-endpoints.

The agents model circular warehouse robots. All agents use the same rotational
velocity $v_{\emph{rot}}$. The following rules impose restrictions on their
legal movements and translational velocities: All free agents can move with high
translational free velocity $v_{\emph{trans}} = v_{\emph{free}}$ through all
cells because warehouse robots that do not carry shelves can move through all
cells, including those that house shelves. All task agents can move with slow
translational task velocity $v_{\emph{trans}} = v_{\emph{task}}$ through only
the pickup and delivery endpoints of their tasks and all other non-endpoints
since warehouse robots that carry shelves cannot move through cells that house
shelves.

\section{Experimental Results}

\begin{table}[t]
\Large
\centering
\caption{Experiment 1. (Inapplicable entries are dashed.)}
\label{tab:1}
\setlength{\tabcolsep}{2.5pt}
\resizebox{\columnwidth}{!}{
\begin{tabular}{c|c||c|c|r|c|r|c|c|c}
\hline
algorithm                         & $v_{\emph{task}}$ & \begin{tabular}[c]{@{}c@{}}discrete\\srvc time\end{tabular} & \begin{tabular}[c]{@{}c@{}}discrete\\makespan\end{tabular} & \multicolumn{1}{c|}{\begin{tabular}[c]{@{}c@{}}srvc\\time\end{tabular}} & \begin{tabular}[c]{@{}c@{}}make\\span\end{tabular} & \multicolumn{1}{c|}{\begin{tabular}[c]{@{}c@{}}plan\\ time\end{tabular}} & \begin{tabular}[c]{@{}c@{}}post-proc\\ time\end{tabular} & thpt & \begin{tabular}[c]{@{}c@{}}stdy\\thpt\end{tabular} \\
\hline \hline
\multirow{3}{*}{\begin{tabular}[c]{@{}l@{}}TP-SIPPwRT\end{tabular}} & 0.50  & \multirow{3}{*}{--}      & \multirow{3}{*}{--}    & 944.03  & 2,475.58 & 0.90                     & \multirow{3}{*}{--} & 0.397 & 0.433 \\
                                                                          & 0.75 &                         &                       & 601.69  & 1,755.22 & 0.92                     &                    & 0.552 & 0.632 \\
                                                                          & 1.00    &                         &                       & 435.26  & 1,392.00 & 0.83                     &                    & 0.689 & 0.782 \\\hline
\multirow{3}{*}{\begin{tabular}[c]{@{}l@{}}CENTRAL\end{tabular}}      & 0.50  & \multirow{3}{*}{325.28} & \multirow{3}{*}{1,163} & 1,049.51 & 2,617.00 & \multirow{3}{*}{1,161.44} & 264.66             & 0.370 & 0.406 \\
                                                                          & 0.75 &                         &                       & 691.90  & 1,895.68 &                          & 254.36             & 0.504 & 0.552 \\
                                                                          & 1.00    &                         &                       & 520.36  & 1,553.00 &                          & 269.91             & 0.609 & 0.670 \\\hline
\multirow{3}{*}{\begin{tabular}[c]{@{}l@{}}TP-A*\end{tabular}}         & 0.50  & \multirow{3}{*}{329.83} & \multirow{3}{*}{1,204} & 1,026.23 & 2,628.22 & \multirow{3}{*}{1.00}    & 267.38             & 0.373 & 0.408 \\
                                                                          & 0.75 &                         &                       & 675.65  & 1,909.45 &                          & 295.54             & 0.508 & 0.558 \\
                                                                          & 1.00    &                         &                       & 505.81  & 1,570.77 &                          & 278.74             & 0.609 & 0.683                                                        \\
\hline
\end{tabular}
}
\end{table}

\begin{figure*}[t]
  \centering
  \includegraphics[width=0.3\textwidth]{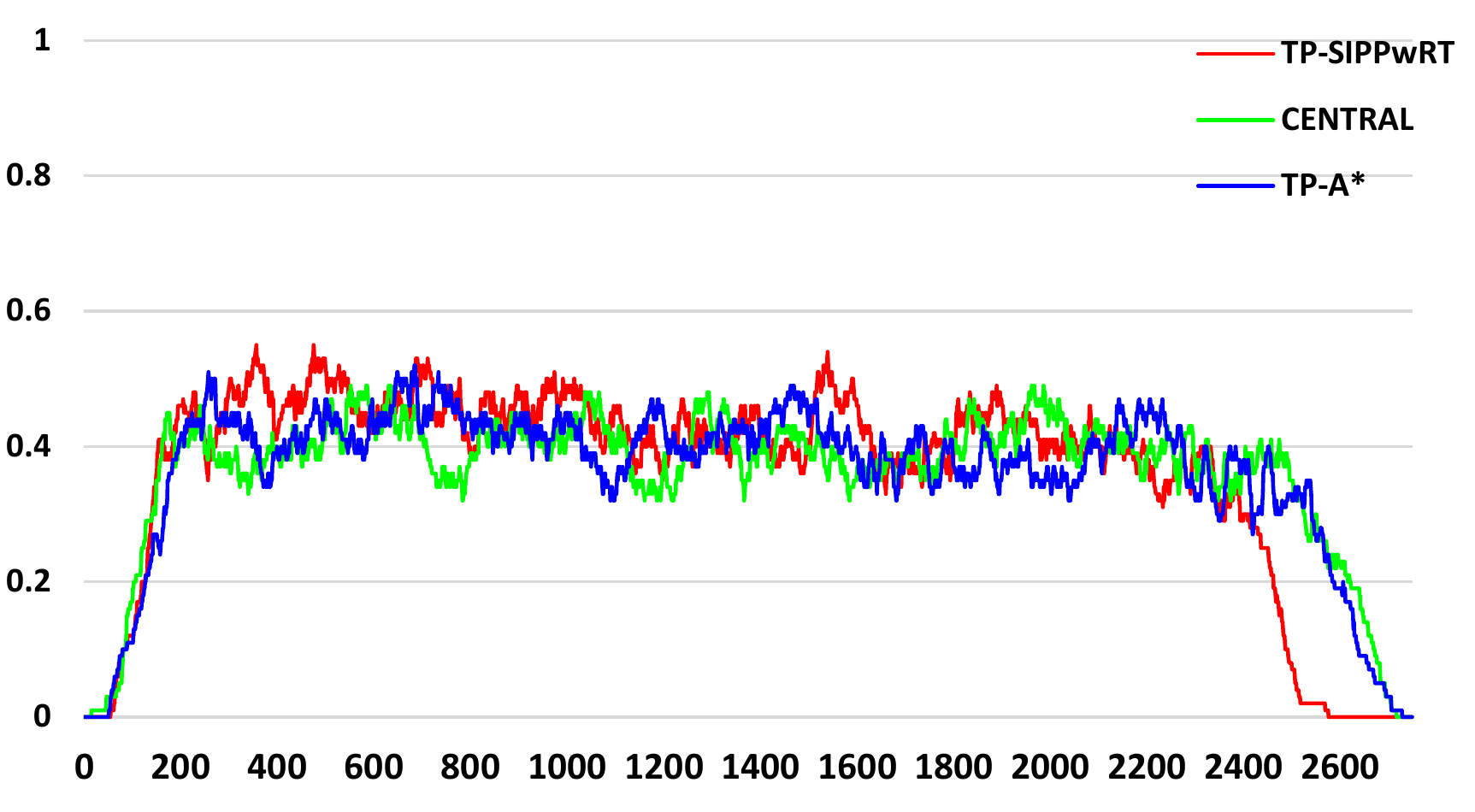}\includegraphics[width=0.3\textwidth]{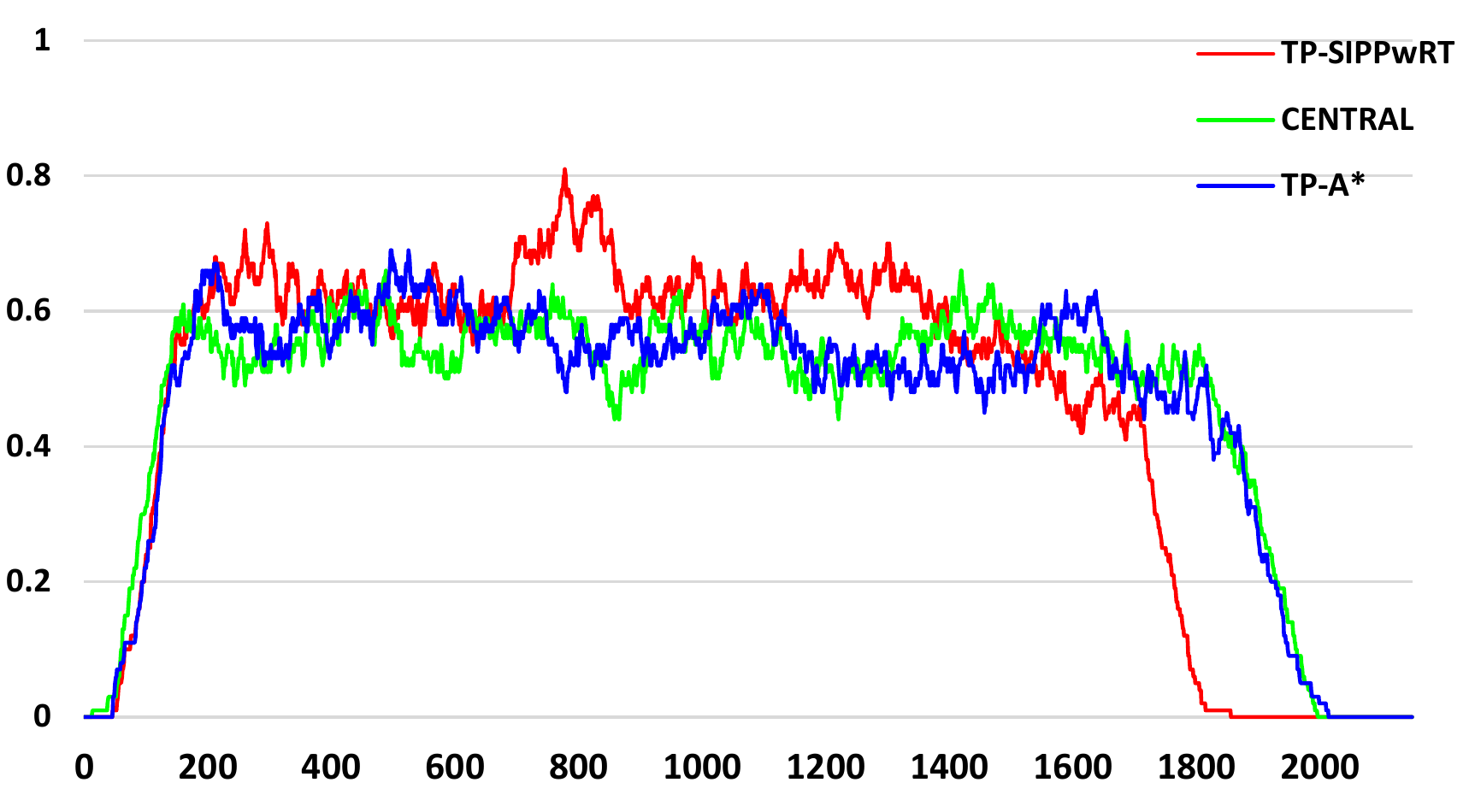}\includegraphics[width=0.3\textwidth]{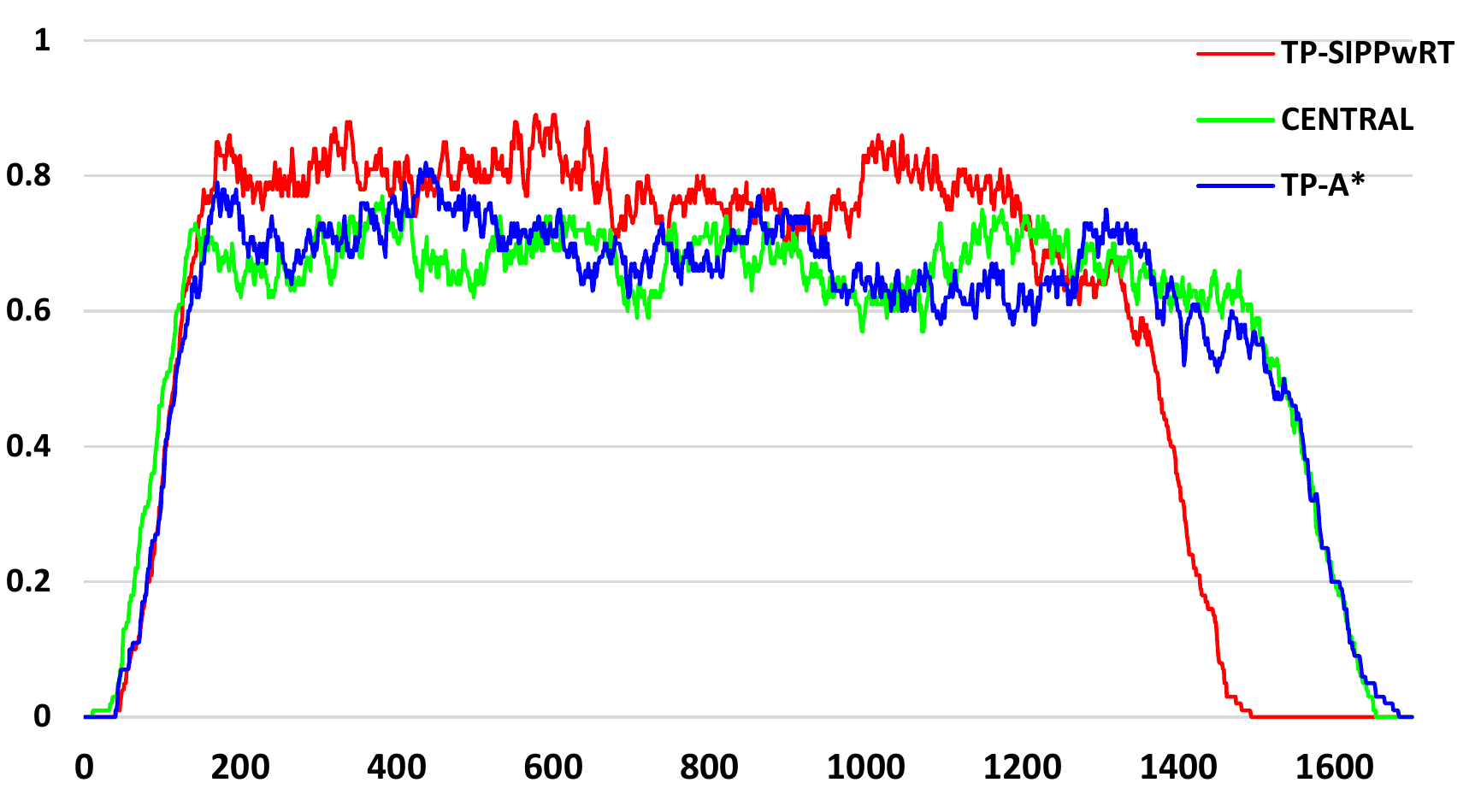}
  \caption{Number of tasks executed per second in a moving 100-second window
    $(t-100,t]$ (that is, throughput at time $t$) as a function of time $t$ for
  different MAPD algorithms. Left: $v_{\emph{task}} = \SI{0.50}{m/s}$. Middle:
  $v_{\emph{task}} = \SI{0.75}{m/s}$. Right: $v_{\emph{task}} =
  \SI{1.00}{m/s}$.}
    \label{fig:three_solver}
\end{figure*}

We now report our experimental results on a 2.50 GHz Intel Core i5-2450M laptop
with 6 GB RAM. Videos of sample experiments can be found at

{\centering
http://idm-lab.org/project-p.html
}

\noindent\textbf{Experiment 1: MAPD Algorithms and Task Velocity.}
We compare TP-SIPPwRT for $v_{\emph{task}}= 0.50$, $0.75$, and
$\SI{1.00}{m/s}$ on the agent simulator in the small simulated warehouse
environment of Figure \ref{fig:warehouse} (left) to two MAPD algorithms that
both assume discrete agent movements with uniform velocity to the four
neighboring cells, namely the original TP \cite{MaAAMAS17} (labeled TP-A*) and CENTRAL
\cite{MaAAMAS17}, which repeatedly uses the Hungarian Method \cite{Kuhn1955}
to assign tasks to agents and then Conflict-Based Search
\cite{DBLP:journals/ai/SharonSFS15} to plan paths for the agents. We first
convert the paths produced by these two MAPD algorithms from containing
movements in the four compass directions to containing forward movements and
point turns.  We then use MAPF-POST \cite{HoenigICAPS16} to adapt the paths in
polynomial time to continuous agent movements with given velocities. Since MAPF-POST guarantees safety distances between agents of
$L/\sqrt{2} = \SI{0.71}{m}$, we use the same radius of $R = 0.5 L/\sqrt{2} =
\SI{0.35}{m}$ for all agents. We use a runtime limit of 5 minutes per
instance. We use 30 agents ({\em agts}) since CENTRAL, the most
runtime-intensive of our MAPD algorithms, can handle only slightly more than
30 agents without any timeouts. We use $v_{\emph{free}} = \SI{1.00}{m/s}$ and
$v_{\emph{rot}} = \pi/2 = \SI{1.57}{rad/s}$. We generate one sequence of 1,000
tasks and insert them in the generated order into the system with a task
frequency ({\it task freq}) of 2 tasks in the beginning of every second.

Figure~\ref{fig:three_solver} visualizes the throughput at time $t$ [number of tasks that finish execution per second in the 100-second window $(t-100,t]$], measured in
  tasks per second, as a function of $t$, measured in seconds. The steady
  state is the time interval when the throughput remains mostly unchanged,
  determined by visual inspection of the graphs. We use as steady state
  $t\in[501,2100]$ for $v_{\emph{task}}=\SI{0.50}{m/s}$, $t\in[501,1500]$ for
  $v_{\emph{task}}=\SI{0.75}{m/s}$, and $t\in[501,1100]$ for
  $v_{\emph{task}}=\SI{1.00}{m/s}$. The throughput at time $t$ of TP-SIPPwRT
  decreases earlier than the ones of TP-A* and CENTRAL because fewer still
  unexecuted tasks are available toward the end for TP-SIPPwRT
  than for them. Thus, TP-SIPPwRT is more effective than them.

Table~\ref{tab:1} reports the discrete service ({\it srvc}) time [time until a
  task has finished execution after insertion into the system according to the original
  plan with discrete agent movements, averaged over all tasks], discrete
makespan [time when the last task has finished execution according to the original plan with
  discrete agent movements], service ({\it srvc}) time [time until a task has
  actually finished execution after insertion into the system, averaged over all tasks], makespan [time when the last
  task has actually finished execution], planning ({\it plan}) time [execution time of
  the MAPD algorithm], and post-processing ({\it post-proc}) time [execution
  time of MAPF-POST], all measured in seconds, as well as the throughput ({\it
  thpt}) [throughput at time $t$ averaged over all times $t$ whose throughputs are positive] and the throughput in the steady state ({\it stdy
  thpt}) [throughput at time $t$ averaged over all times in the steady state],
both measured in number of tasks per second. Service time, makespan, and
throughput measure effectiveness, while the planning and post-processing times
measure efficiency. The planning time of TP-SIPPwRT is less than one second
for 30 agents and 1,000 tasks. It is on par with the one of TP-A* and smaller
than the one of CENTRAL. Furthermore, TP-SIPPwRT does not have any
post-processing time while both TP-A* and CENTRAL have post-processing times
of more than 250 seconds. Thus, TP-SIPPwRT is more efficient than them. The
service time and makespan of TP-SIPPwRT are smaller than the ones of TP-A* and
CENTRAL, while its throughput is larger.  Thus, TP-SIPPwRT is more effective
than them.

\begin{table}[t]
\Large
\centering
\caption{Experiment 2.}
\label{tab:2}
\setlength{\tabcolsep}{2.5pt}
\resizebox{\columnwidth}{!}{
\begin{tabular}{c|c||r|r|r|r|r|r|r|r|r|r|r|r}
\hline
\multicolumn{2}{c||}{$v_{task}$}     & \multicolumn{4}{c|}{0.50}                                                                                                                                        & \multicolumn{4}{c|}{0.75}                                                                                                                                       & \multicolumn{4}{c}{1.00}                                                                                                                                          \\
\hline
\multicolumn{1}{c|}{agts}
& \multicolumn{1}{c||}{\begin{tabular}[c]{@{}c@{}}task\\ freq\end{tabular}}
& \multicolumn{1}{c|}{\begin{tabular}[c]{@{}c@{}}srvc\\ time \end{tabular}}
& \multicolumn{1}{c|}{\begin{tabular}[c]{@{}c@{}}make\\ span\end{tabular}}
& \multicolumn{1}{c|}{\begin{tabular}[c]{@{}c@{}}plan\\time \end{tabular}}
& \multicolumn{1}{c|}{\begin{tabular}[c]{@{}c@{}}thpt\end{tabular}}
& \multicolumn{1}{c|}{\begin{tabular}[c]{@{}c@{}}srvc\\ time\end{tabular}}
& \multicolumn{1}{c|}{\begin{tabular}[c]{@{}c@{}}make\\ span\end{tabular}}
& \multicolumn{1}{c|}{\begin{tabular}[c]{@{}c@{}}plan\\time \end{tabular}}
& \multicolumn{1}{c|}{\begin{tabular}[c]{@{}c@{}}thpt\end{tabular}}
& \multicolumn{1}{c|}{\begin{tabular}[c]{@{}c@{}}srvc\\ time\end{tabular}}
& \multicolumn{1}{c|}{\begin{tabular}[c]{@{}c@{}}make\\ span\end{tabular}}
& \multicolumn{1}{c|}{\begin{tabular}[c]{@{}c@{}}plan\\time \end{tabular}}
& \multicolumn{1}{c}{\begin{tabular}[c]{@{}c@{}}thpt\end{tabular}} \\
\hline \hline
\multirow{4}{*}{10} & 1  & 2,809.72 & 6,771.00 & 0.84 & 0.146 & 1,834.97 & 4,764.28 & 0.86 & 0.213 & 1,357.21 & 3,818.00 & 0.72 & 0.270 \\
                    & 2  & 3,029.59 & 6,759.41 & 0.85 & 0.157 & 2,077.68 & 4,768.89 & 0.84 & 0.215 & 1,584.62 & 3,784.00 & 0.73 & 0.274 \\
                    & 5  & 3,181.97 & 6,789.41 & 0.86 & 0.155 & 2,185.29 & 4,748.33 & 0.86 & 0.225 & 1,710.76 & 3763.71 & 0.75 & 0.274 \\
                    & 10 & 3,215.43 & 6,775.00 & 0.84 & 0.159 & 2,252.70 & 4,762.45 & 0.88 & 0.219 & 1,750.19 & 3,749.00 & 0.75 & 0.280 \\\hline
\multirow{4}{*}{20} & 1  & 1,228.35 & 3,557.58 & 0.90 & 0.295 & 745.48  & 2,540.33 & 0.89 & 0.411 & 502.50  & 2,000.71 & 0.76 & 0.511 \\
                    & 2  & 1,450.40 & 3,503.00 & 0.89 & 0.298 & 966.27  & 2,493.67 & 0.91 & 0.392 & 714.20  & 1,980.00 & 0.79 & 0.489 \\
                    & 5  & 1,591.79 & 3,519.83 & 0.89 & 0.292 & 1,088.86 & 2,481.85 & 0.88 & 0.416 & 844.32  & 1,966.00 & 0.81 & 0.507 \\
                    & 10 & 1,661.62 & 3,502.83 & 0.88 & 0.290 & 1,136.08 & 2,479.45 & 0.90 & 0.417 & 892.22  & 1,964.00 & 0.81 & 0.507 \\\hline
\multirow{4}{*}{30} & 1  & 723.03  & 2,482.41 & 0.94 & 0.396 & 389.15  & 1,763.50 & 0.91 & 0.551 & 222.21  & 1,431.71 & 0.82 & 0.672 \\
                    & 2  & 944.03  & 2,475.58 & 0.90 & 0.397 & 601.69  & 1,755.22 & 0.92 & 0.552 & 435.26  & 1,392.00 & 0.83 & 0.689 \\
                    & 5  & 1,079.62 & 2,435.83 & 0.90 & 0.398 & 728.33  & 1,724.18 & 0.92 & 0.555 & 563.14  & 1,372.71 & 0.83 & 0.688 \\
                    & 10 & 1,126.47 & 2,468.00 & 0.93 & 0.393 & 779.67  & 1,737.00 & 0.92 & 0.550 & 612.06  & 1,380.00 & 0.84 & 0.065 \\\hline
\multirow{4}{*}{40} & 1  & 484.93  & 2,023.58 & 0.90 & 0.484 & 225.18  & 1,471.12 & 0.95 & 0.657 & 101.16  & 1,252.00 & 0.85 & 0.765 \\
                    & 2  & 701.23  & 1,945.00 & 0.94 & 0.503 & 432.11  & 1,430.33 & 0.95 & 0.674 & 298.04  & 1,122.71 & 0.89 & 0.847 \\
                    & 5  & 830.73  & 2,054.00 & 0.89 & 0.470 & 563.25  & 1,368.67 & 0.94 & 0.693 & 427.23  & 1,073.00 & 0.87 & 0.870 \\
                    & 10 & 880.46  & 1,905.00 & 0.88 & 0.506 & 605.10  & 1,382.67 & 0.94 & 0.686 & 469.92  & 1,095.71 & 0.89 & 0.853 \\\hline
\multirow{4}{*}{50} & 1  & 331.66  & 1,680.41 & 0.98 & 0.641 & 122.98  & 1,262.00 & 0.98 & 0.771 & 63.45   & 1,140.41 & 0.96 & 0.845 \\
                    & 2  & 557.10  & 1,676.58 & 0.97 & 0.581 & 335.58  & 1,192.51 & 0.96 & 0.804 & 219.99  & 968.00  & 0.92 & 0.976 \\
                    & 5  & 683.56  & 1,674.41 & 0.97 & 0.573 & 454.42  & 1,153.51 & 0.93 & 0.814 & 344.44  & 931.00  & 0.91 & 0.992 \\
                    & 10 & 729.07  & 1,644.41 & 0.97 & 0.582 & 502.03  & 1,200.94 & 0.98 & 0.784 & 389.78  & 926.00  & 0.93 & 0.996 \\
                    \hline
\end{tabular}
}
\end{table}

\noindent\textbf{Experiment 2: Number of Agents, Task Frequency, and Task Velocity.}
We run TP-SIPPwRT with the same setup as in Experiment 1 (including the same
sequence of 1,000 tasks) for $v_{\emph{task}}= 0.50$, $0.75$, and
$\SI{1.00}{m/s}$, 10, 20, 30, 40, and 50 agents, and task frequencies of 1, 2,
5, and 10 tasks per second. Table~\ref{tab:2} shows that the planning time of
TP-SIPPwRT is less than one second for up to 50 agents and 1,000 tasks.
As expected, the service time decreases as the task frequency decreases; the service time and makespan decrease and the throughput increases as the number of agents increases; and the service time and makespan decrease and the throughput increases as the task velocity increases.


\begin{table}[t]
\Large
\centering
\caption{Experiment 3.}
\label{tab:3}
\setlength{\tabcolsep}{2.5pt}
\resizebox{\columnwidth}{!}{
\begin{tabular}
{r||r|r|r|r|r|r|r|r|r|r|r|r|r|r|r}
\hline
$v_{task}$    & \multicolumn{5}{c}{0.50}
& \multicolumn{5}{|c}{0.75}
& \multicolumn{5}{|c}{1.00}
\\ \hline
\multicolumn{1}{c||}{\begin{tabular}[c]{@{}c@{}}agts\end{tabular}}
& \multicolumn{1}{c|}{\begin{tabular}[c]{@{}c@{}}srvc\\ time\end{tabular}}
& \multicolumn{1}{c|}{\begin{tabular}[c]{@{}c@{}}make\\ span\end{tabular}}
& \multicolumn{1}{c|}{\begin{tabular}[c]{@{}c@{}}plan\\ time \end{tabular}}
& \multicolumn{1}{c|}{thpt}
& \multicolumn{1}{c|}{\begin{tabular}[c]{@{}c@{}}stdy\\ thpt\end{tabular}}
& \multicolumn{1}{c|}{\begin{tabular}[c]{@{}c@{}}srvc\\ time\end{tabular}}
& \multicolumn{1}{c|}{\begin{tabular}[c]{@{}c@{}}make\\ span\end{tabular}}
& \multicolumn{1}{c|}{\begin{tabular}[c]{@{}c@{}}plan\\ time \end{tabular}}
& \multicolumn{1}{c|}{thpt}
& \multicolumn{1}{c|}{\begin{tabular}[c]{@{}c@{}}stdy\\ thpt\end{tabular}}
& \multicolumn{1}{c|}{\begin{tabular}[c]{@{}c@{}}srvc\\ time\end{tabular}}
& \multicolumn{1}{c|}{\begin{tabular}[c]{@{}c@{}}make\\ span\end{tabular}}
& \multicolumn{1}{c|}{\begin{tabular}[c]{@{}c@{}}plan\\ time \end{tabular}}
& \multicolumn{1}{c|}{thpt}
& \multicolumn{1}{c}{\begin{tabular}[c]{@{}c@{}}stdy\\ thpt\end{tabular}} \\
\hline \hline
100 & 877.94 & 2,891.58 & 5.72 & 0.70 & 0.81 & 489.46 & 2,130.67 & 5.83  & 0.91 & 1.15 & 289.80 & 1,671.00 & 5.15  & 1.16 & 1.49 \\\hline
150 & 525.07 & 2,269.58 & 6.49 & 0.88 & 1.15 & 253.49 & 1,602.00 & 6.67  & 1.20 & 1.64 & 122.69 & 1,396.71 & 5.57  & 1.37 & 1.98 \\\hline
200 & 353.46 & 1,905.58 & 7.08 & 1.03 & 1.47 & 154.76 & 1,504.63 & 7.35  & 1.31 & 1.97 & 117.21 & 1,276.12 & 9.50  & 1.50 & 2.04 \\\hline
250 & 267.07 & 1,762.24 & 9.35 & 1.13 & 1.71 & 147.90 & 1,271.67 & 12.83 & 1.52 & 1.99 & 132.45 & 1,297.00 & 15.76 & 1.48 & 2.02
\\\hline
\end{tabular}
}
\end{table}

\begin{figure*}[t]
  \centering
  \includegraphics[width=0.3\textwidth]{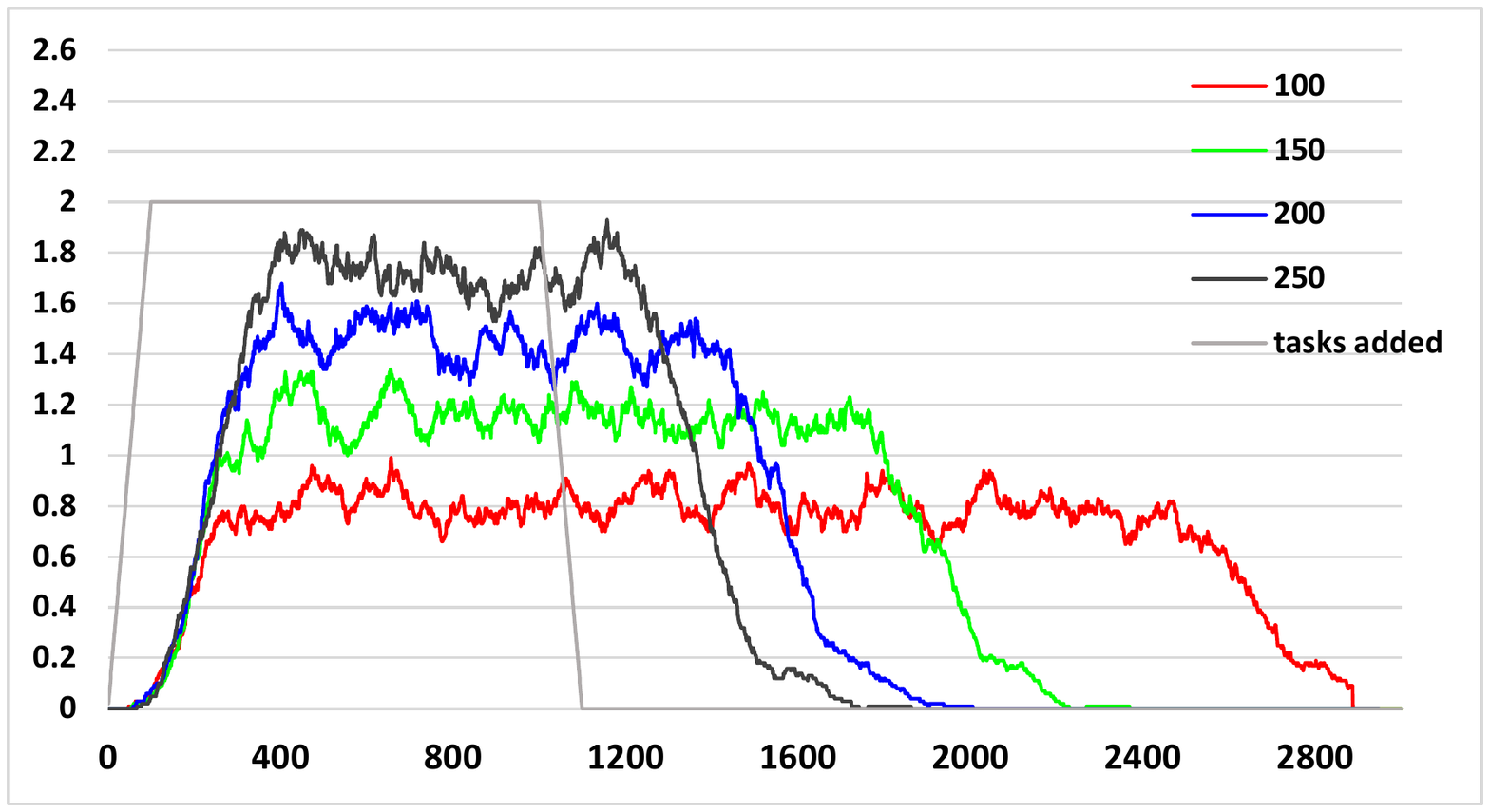}\includegraphics[width=0.3\textwidth]{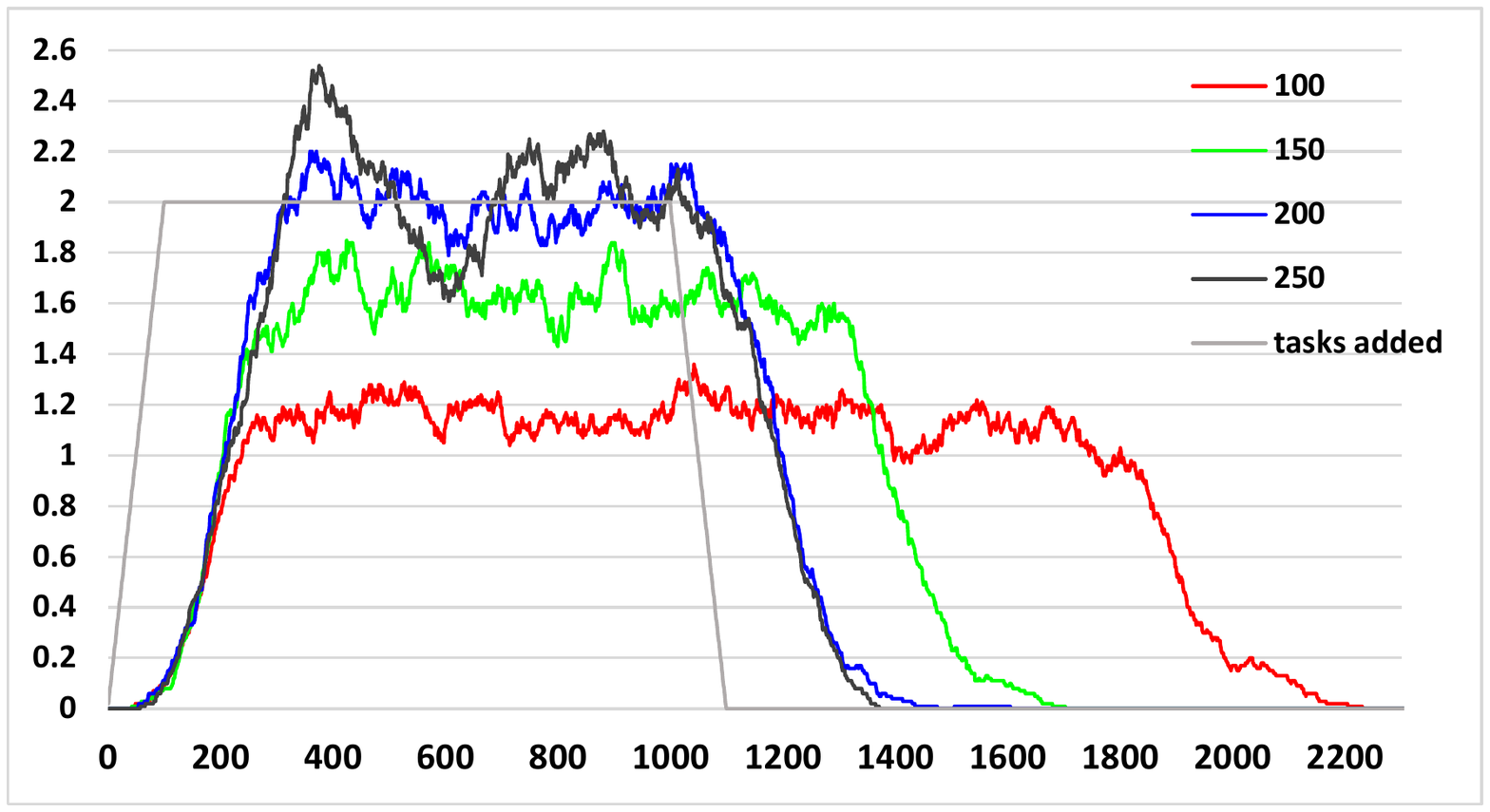}\includegraphics[width=0.3\textwidth]{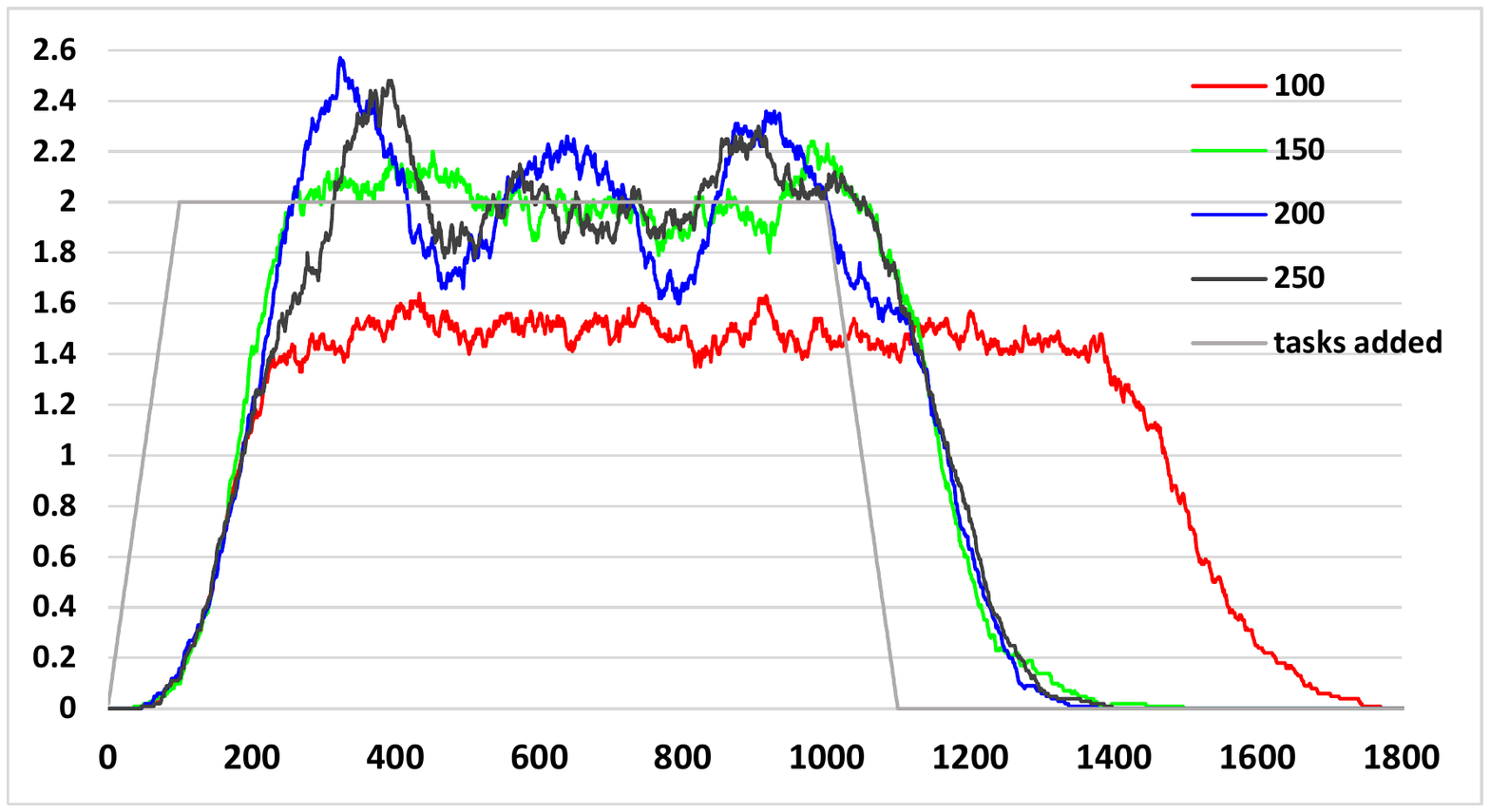}
  \caption{Number of tasks inserted (in light grey) and executed per second in a
    moving 100-second window $(t-100,t]$ as a function of time $t$ for different
  numbers of agents. Left: $v_{\emph{task}} = \SI{0.50}{m/s}$. Middle:
  $v_{\emph{task}} = \SI{0.75}{m/s}$. Right: $v_{\emph{task}} =
  \SI{1.00}{m/s}$.}
  \label{fig:large_results}
\end{figure*}

\noindent\textbf{Experiment 3: Environment Size, Number of Agents, and Task Velocity.}
We run TP-SIPPwRT with the same setup as in Experiment 1 but in the large
simulated warehouse environment of Figure \ref{fig:warehouse} (right) for 100,
150, 200, and 250 agents and $v_{\emph{task}}= 0.50$, $0.75$, and
$\SI{1.00}{m/s}$. We use one sequence of 2,000 tasks and a task frequency of 2
tasks per second. Figure~\ref{fig:large_results} visualizes the throughput at
time $t$. We use as steady state $t\in[501,1000]$.
Table~\ref{tab:3} shows that the planning time of TP-SIPPwRT is less than 16
seconds for up to 250 agents and 2,000 tasks, justifying our claim that it can compute paths for hundreds of agents and thousands of tasks in seconds.
Similarly to before, the service time and makespan decrease and the throughput and planning time increase as the number of agents increases; and the service time and makespan decrease and the throughput increases as the task velocity increases. There is an exception due to the congestion resulting from many agents for 250 agents and $v_{\emph{task}}=\SI{1.00}{m/s}$.

\begin{figure}[t]
  \centering
  \includegraphics[height=60pt]{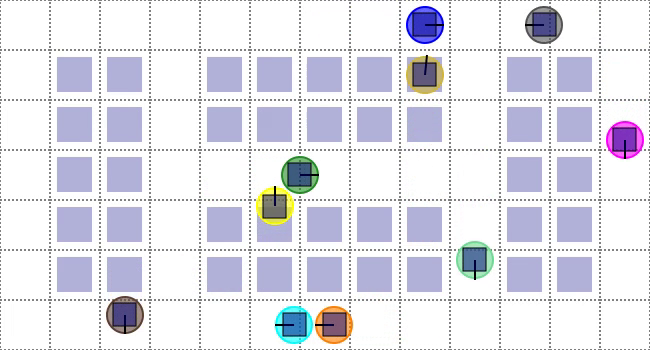}\includegraphics[height=60pt]{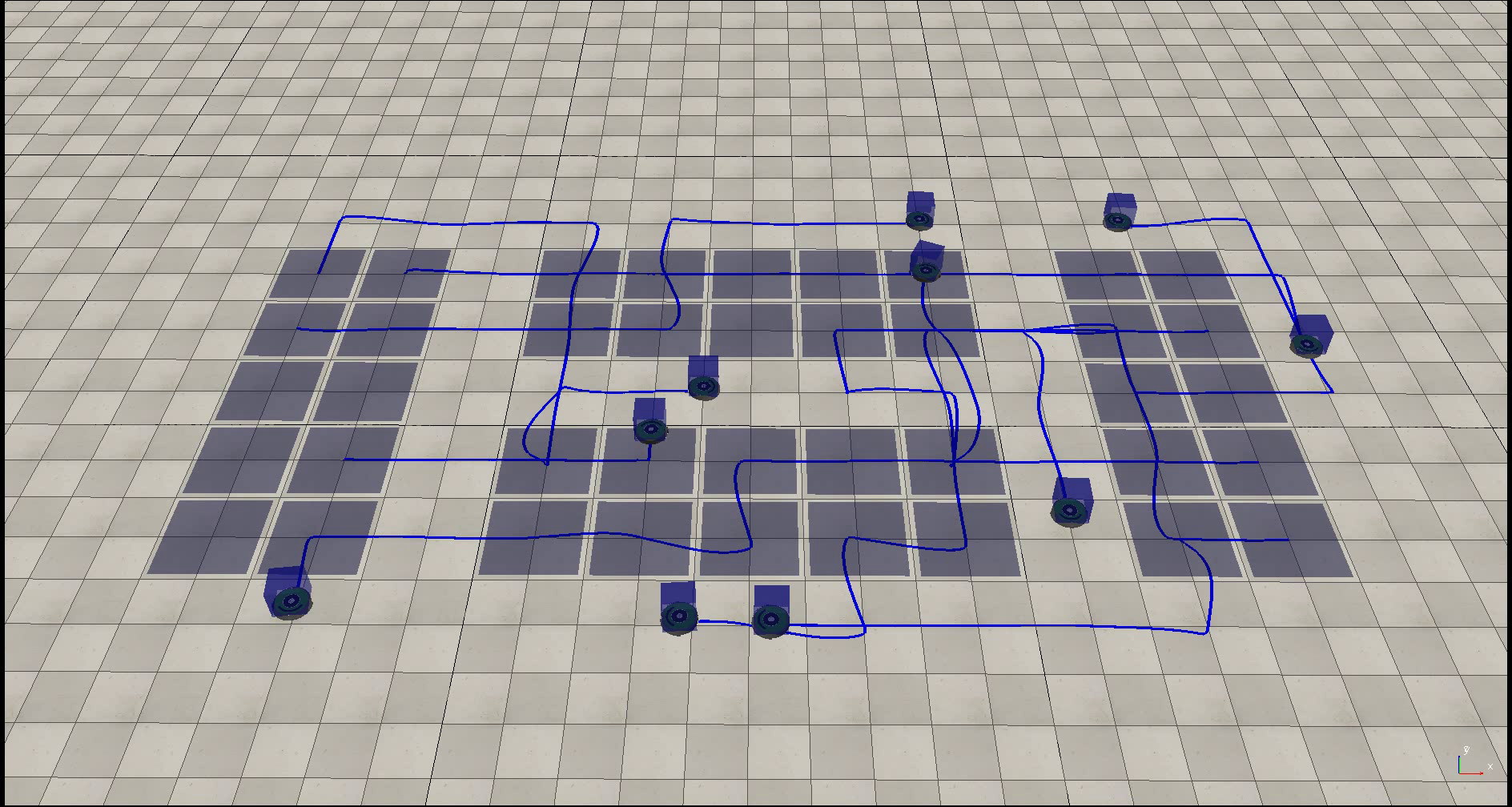}
  \caption{Screenshots for Experiment 3 at $t=\SI{35}{s}$. Left: Agent
    simulator. Right: Robot simulator.}\label{fig:robotics}
\end{figure}

\noindent\textbf{Experiment 4: Robot Simulator.}
We created a custom model of the kinodynamic constraints of a
differential-drive Create2 robot from iRobot for the robot simulator V-REP
\cite{vrep}. Create2 robots have a cylindrical shape with radius \SI{0.175}{m}
and can reach a translational speed of \SI{0.5}{m/s} and a rotational speed
of \SI{4.2}{rad/s}. We use $v_{\emph{free}} = \SI{0.40}{m/s}$,
$v_{\emph{task}} = \SI{0.20}{m/s}$, $v_{\emph{rot}} = \pi = \SI{3.14}{rad/s}$,
and $R = \SI{0.40}{m}$ as conservative values to allow the robots to follow
their paths safely despite unmodeled kinodynamic constraints and motion noise
by TP-SIPPwRT. We implemented a PID controller that uses $[x,y,\theta]^T$
(given by V-REP) as the current state and the desired next cell with the
associated desired arrival time (given by TP-SIPPwRT) as the goal state. The
PID controller corrects for heading errors by orienting the robot to face the
desired next cell while simultaneously adjusting the translational speed to
let the robot arrive at the desired next cell at the desired arrival time. We
limit our experiment to the small warehouse environment of Figure
\ref{fig:robotics} for 10 robots due to the slow runtime of V-REP. We use one
sequence of 20 tasks and a task frequency of 2 tasks per second. The planning
time of TP-SIPPwRT is \SI{2}{ms}. All robots follow their paths safely,
resulting in a service time of \SI{90.57}{s} and a makespan of \SI{171.16}{s}.

\section{Conclusions and Future Work}

We presented the efficient and effective algorithm TP-SIPPwRT for the
Multi-Agent Pickup and Delivery problem. We suggest the following future
research directions: (1) Make existing (even optimal) multi-agent pathfinding
algorithms more general by combining them with our SIPPwRT to compute
continuous agent movements with given velocities. The resulting algorithms
could, for example, be used to make CENTRAL more general. (2) Include
additional kinodynamic constraints into SIPPwRT and TP-SIPPwRT, such as
acceleration and deceleration constraints, to allow robots to follow their
paths even more safely. (3) Make TP-SIPPwRT decentralized.

\small
\bibliographystyle{aaai}
\bibliography{references}

\end{document}


\title{Supplementary Material for:\\Lifelong Path Planning with Kinematic Constraints\\for Multi-Agent Pickup
  and Delivery
}


\maketitle

\section{Proof of Theorem 1}

\begin{pf}[\emph{By Induction}]
Consider a path returned by SIPPwRT. By definition, the agent starts in the first configuration $n_\emph{start}.\emph{cfg}$ at time $\emph{current\_t}$ without any collisions. Assume that the agent arrives at cell $n.\emph{cfg}.\emph{cell}$ with configuration $n.\emph{cfg}$ all the way from the first configuration $n_\emph{start}.\emph{cfg}$ without any collisions. Let $n'.\emph{cfg}$ be the (successor) configuration next to configuration $n.\emph{cfg}$ in the path. As the agent moves from cell $l = n.\emph{cfg}.\emph{cell}$ to the next cell $l' = n'.\emph{cfg}.\emph{cell}$, we use the following arguments:
\begin{enumerate}[(1)]
  \item Line \ref{lin:lb2} (\ref{lin:ub2}) (when node $n$ is being expanded) guarantees that any dynamic obstacle, which departs from cell $l$ at a time earlier (later) than when the agent departs from cell $l$ and which then moves in the same direction as the agent, must arrive at the next cell $l'$ earlier (later) than when the agent arrives at cell $l'$.
  \item Line \ref{lin:ub1} (when node $n$ is being expanded) guarantees that the agent must not collide with any dynamic obstacle, which arrives at cell $l$ in a non-opposite direction at a time later than when the agent departs from cell $l$, until the dynamic obstacle has arrived at cell $l$.
  \item Line \ref{lin:lb1}  (when node $n$ is being expanded and node $n'$ is being generated) guarantees that the agent must not collide with any dynamic obstacle, which departs from cell $l'$ in a non-opposite direction at a time earlier than when the agent arrives at cell $l'$, until the agent has arrived at cell $l'$. The line also guarantees that, if any dynamic obstacle departs from cell $l'$ in the opposite direction at a time earlier than when the agent arrives at cell $l'$, it also arrives at cell $l$ at a time earlier than when the agent departs from cell $l$ and thus also departs from cell $l$ at a time earlier than when the agent arrives at cell $l$ because its corresponding reserved interval does not intersect with the safe interval $n.\emph{int}$ for cell $l$. Therefore, in both cases, the agent does not collide with such a dynamic obstacle (that departs from cell $l'$ at a time earlier than when the agent arrives at cell $l'$) as it moves from cell $l$ to cell $l'$ due to the induction assumption.
  \item For a non-goal node $n'$, Line \ref{lin:ub1} (when node $n'$ is being expanded) guarantees that the agent must not collide with any dynamic obstacle, which arrives at cell $l'$ in a non-opposite direction at a time later than when the agent departs from cell $l'$, until the dynamic obstacle has arrived at cell $l'$.
\end{enumerate}
From the induction assumption, it suffices to prove that the agent does not collide with any given dynamic obstacle as the agent departs from cell $l$ until it arrives at cell $l'$. Assume, for a proof by contradiction, that the agent collides with the dynamic obstacle.
\begin{enumerate}
\item[(i)] If the dynamic obstacle arrives at cell $l$ at a time earlier than when the agent departs from cell $l$, it also departs from cell $l$ at a time earlier than when the agent departs from cell $l$ because its corresponding reserved interval does not intersect with the safe interval $n.\emph{int}$ for cell $l$. It can collide with the agent only if it then moves from cell $l$ to cell $l'$ at a fixed velocity $v_\emph{trans}'$ smaller than the fixed velocity $v_\emph{trans}$ at which the agent moves from cell $l$ to cell $l'$. It must arrive at cell $l'$ at a time earlier than when the agent arrives at cell $l'$ due to (1) and also departs at a time earlier than when the agent arrives at cell $l'$ because its corresponding reserved interval does not intersect with the safe interval $n'.\emph{int}$ for cell $l'$. The collision thus remains when the agent arrives at cell $l'$, which contradicts (3).
\item[(ii.a)] If the dynamic obstacle arrives at cell $l$ in a non-opposite direction at a time later than when the agent departs from cell $l$, it does not collide with the agent until it arrives at cell $l$ due to (2). As a consequence, the dynamic obstacle also departs from cell $l$ at a time later than when the agent departs from cell $l$. It can collide with the agent only if it then moves from cell $l$ to cell $l'$ at a fixed velocity $v_\emph{trans}'$ larger than the fixed velocity $v_\emph{trans}$ at which the agent moves from cell $l$ to cell $l'$. It must arrive at cell $l'$ at a time later than when the agent arrives at cell $l'$ due to (1) and also later than when the agent departs from cell $l'$ because its corresponding reserved interval does not intersect with the safe interval $n'.\emph{int}$ for cell $l'$. Node $n'$ is thus a non-goal node due to Line \ref{lin:is_goal} (after $n'$ is popped from \emph{OPEN}), and the collision thus remains when the agent departs from cell $l'$, which contradicts (4).
\item[(ii.b)] If the dynamic obstacle arrives at cell $l$ in the opposite direction at a time later than when the agent departs from cell $l$, Line \ref{lin:ub1} (when node $n$ is being expanded) guarantees that it departs form cell $l'$ at a time later than when the agent arrives at cell $l'$. The dynamic obstacle thus also arrives at cell $l'$ at a time later than when the agent departs from cell $l'$ because its corresponding reserved interval does not intersect with the safe interval $n'.\emph{int}$ for cell $l'$. Node $n'$ is thus a non-goal node due to Line \ref{lin:is_goal} (after $n'$ is popped from \emph{OPEN}). If the dynamic obstacle arrives at cell $l'$ from another cell $l''$ in the opposite direction to the one in which the agent departs from cell $l'$ (toward cell $l''$), Line \ref{lin:ub1} (Case (c) of Function $\emph{Offset}$) (when node $n'$ is being expanded) guarantees that the dynamic obstacle arrives at cell $l''$ at a time later than when the agent departs from cell $l'$, ensuring that the agent does not collide with the dynamic obstacle until the agent departs from cell $l'$. Otherwise, any collision thus remains when the agent departs from cell $l'$, which contradicts (4).
\item[(iii)] If the dynamic obstacle does not fall into Case (i) but also arrives at cell $l'$ at a time earlier than when the agent arrives at cell $l'$, it must also depart from cell $l'$ earlier than when the agent arrives at cell $l'$ because its corresponding reserved interval does not intersect with the safe interval $n'.\emph{int}$ for cell $l'$. Any collision thus remains when the agent arrives at cell $l'$, which contradicts (3).
\item[(iv.a)] If node $n'$ is a non-goal node and the dynamic obstacle does not fall into Case (ii.b) but also arrives at cell $l'$ at a time later than when the agent arrives at cell $l'$, the dynamic obstacle must arrives at cell $l'$ later than when the agent departs from cell $l'$ because its corresponding reserved interval does not intersect with the safe interval $n'.\emph{int}$ for cell $l'$. We use the same argument as for Case (ii.b): If the dynamic obstacle arrives at cell $l'$ from another cell $l''$ in the opposite direction to the one in which the agent departs from cell $l'$ (toward cell $l''$), Line \ref{lin:ub1} (Case (c) of Function $\emph{Offset}$) (when node $n'$ is being expanded) guarantees that the dynamic obstacle arrives at cell $l''$ at a time later than when the agent departs from cell $l'$, ensuring that the agent does not collide with the dynamic obstacle until the agent departs from cell $l'$. Otherwise, any collision thus remains when the agent departs from cell $l'$, which contradicts (4).
\item[(iv.b)] If node $n'$ is the goal node $n_\emph{goal}$, no dynamic obstacle arrives at cell $l'$ at a time later than when the agent arrives at cell $l'$ due to Line \ref{lin:is_goal} (after $n'$ is popped from \emph{OPEN}). No collision is possible until the agent has arrived at cell $l'$.
\end{enumerate}
Therefore, the agent arrives at cell $l' = n'.\emph{cfg}.\emph{cell}$ with configuration $n'.\emph{cfg}$ without any collisions.
\end{pf}